\documentclass{article}
\usepackage{aikyam}						

\usepackage{booktabs}						
\usepackage{multirow}						
\usepackage{amsfonts}						
\usepackage{graphicx}						
\usepackage{duckuments}						
\usepackage{amssymb}
\usepackage{bbm}

\usepackage[numbers,compress,sort]{natbib}	
\usepackage{tikz}
\usepackage{float}
\usetikzlibrary{arrows.meta, positioning}
\usepackage{microtype}

\title[Quantifying Explanation Quality in Graph Neural Networks using Out-of-Distribution Generalization]{\LARGE \centering Quantifying Explanation Quality in Graph Neural Networks using Out-of-Distribution Generalization}

\author[Last et al.]{%
Ding Zhang\\
University of Virginia \\\And
Siddharth Betala\\
Entalpic \\\And
Chirag Agarwal\\
University of Virginia
}

\newcommand{\xhdr}[1]{\vspace{0em}\noindent{{\bf #1.}}}
\newcommand{\ie}{\textit{i.e., \xspace}}
\newcommand{\eg}{\textit{e.g., \xspace}}

\newcommand{\hide}[1]{}

\newcommand{\std}[1]{\scriptsize{$\pm$#1}}

\usepackage{colortbl}

\definecolor{Gray}{gray}{0.9}
\definecolor{LightCyan}{rgb}{0.88,1,1}
\definecolor{darkred}{rgb}{0.8,0.1,0.1}
\definecolor{darkyellow}{rgb}{0.95, 0.68, 0.22}
\definecolor{darkgreen}{rgb}{0.1,0.8,0.1}
\newcolumntype{a}{>{\columncolor{Gray}}c}
\newcolumntype{b}{>{\columncolor{white}}c}

\newcommand{\fc}{\textsc{Fc}\xspace}
\newcommand{\fluoride}{\textsc{Fluoride Carbonyl}\xspace}
\newcommand{\mutag}{\textsc{Mutag}\xspace}

\newcommand{\egs}{\textsc{Egs}\xspace}
\newcommand{\tri}{\textsc{Triangle}\xspace}
\newcommand{\pen}{\textsc{Pentagon}\xspace}

\begin{document}

\maketitle
\begin{abstract}
    Evaluating the quality of post-hoc explanations for Graph Neural Networks (GNNs) remains a significant challenge. While recent years have seen an increasing development of explainability methods, current evaluation metrics (\eg fidelity, sparsity) often fail to assess whether an explanation identifies the true underlying causal variables. To address this, we propose the \textbf{E}xplanation-\textbf{G}eneralization \textbf{S}core (\egs), a metric that quantifies the causal relevance of GNN explanations. \egs is founded on the principle of feature invariance and posits that if an explanation captures true causal drivers, it should lead to stable predictions across distribution shifts. To quantify this, we introduce a framework that trains GNNs using explanatory subgraphs and evaluates their performance in Out-of-Distribution (OOD) settings (here, OOD generalization serves as a rigorous proxy for the explanation's causal validity). Through large-scale validation involving 11,200 model combinations across synthetic and real-world datasets, our results demonstrate that \egs provides a principled benchmark for ranking explainers based on their ability to capture causal substructures, offering a robust alternative to traditional fidelity-based metrics.    
\end{abstract}

\section{Introduction}
\label{sec:intro}

\begin{figure*}[ht]
\vskip 0.2in
\begin{center}
\centerline{\includegraphics[width=0.9\textwidth]{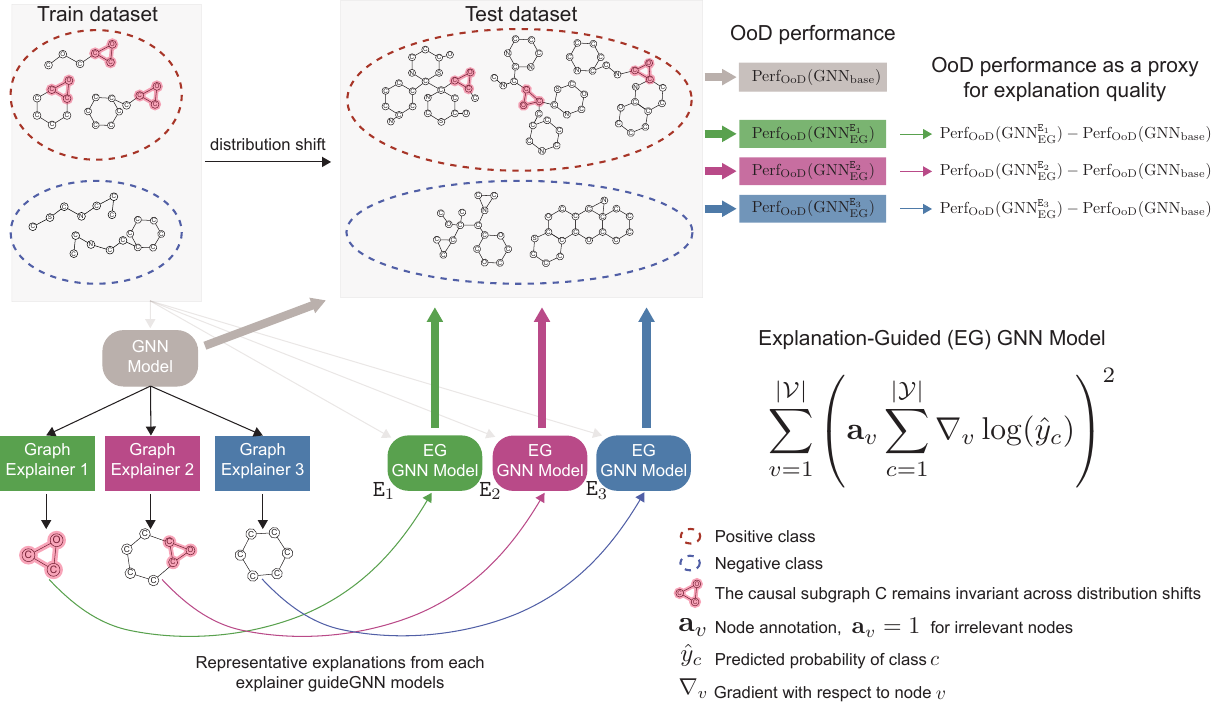}}
\caption{\textbf{Overview of our framework and proposed evaluation metric.} We train GNN models on datasets with a distribution shift between train and test sets and generate explanations using different graph explainers. These explanations are then used to constrain new Explanation-guided (EG) GNN models during training. The resulting out-of-distribution (OOD) performance of EC-GNNs, compared to a vanilla GNN, serves as a proxy for assessing the causal validity of the explanations. The right panel illustrates the formulation of the model optimization constraint, where the gradients of irrelevant nodes are minimized. This methodology enables the evaluation of explanation quality without relying on ground truth explanations.
}
\label{fig:1}
\end{center}
\vskip -0.2in
\end{figure*} 

\looseness=-1 Explainability in Graph Neural Networks (GNNs) is essential for building trust, improving hybrid models, and enabling scientific discovery \cite{2024dai,2018holzinger,2020roscher,2021burkart}. Providing explanations helps domain experts, such as biologists and chemists, validate model decisions and prioritize hypotheses. While there has been a plethora of work introducing methods to explain GNN behavior \citep{yuan2022explainability}, evaluating the quality of these explanations remains a major challenge.


Current evaluation protocols are insufficient for high-stakes scientific applications. In fields like drug discovery, reliable explanations must identify the true structural drivers of a chemical property, such as an antibiotic pharmacophore, rather than merely highlighting correlated artifacts~\cite{wong2024}. However, commonly used metrics (\eg fidelity, sparsity, stability) assess the consistency of an explanation rather than its causal validity. For instance, fidelity measures how well an explanation mimics the model's predictions, but a high fidelity score does not guarantee that the explanation reflects the true data generating process. Similarly, an explanation can be stable across instances yet remain non-causal. Consequently, existing metrics often fail to distinguish between spurious correlations and true causal mechanisms, highlighting an urgent need for evaluation frameworks that explicitly account for causality.
\looseness=-1\xhdr{Present work} In this work, we bridge this gap by linking explainability to the principle of feature invariance. We propose that if an explanation truly captures causal features, relying on it should enable the model to generalize across distribution shifts (see Sec.~\ref{sec:OOD_gen_graphs}). We approach this in two steps: 1) We first demonstrate that guiding the optimization of GNNs with ground-truth causal explanations significantly enhances their Out-of-Distribution (OOD) generalization. This validates the premise that causal features are invariant predictors; and 2) Leveraging this insight, we introduce the \textit{Explanation Generalization Score (\egs)}. We \textbf{invert} the training relationship: instead of using the model to generate explanations, we constrain the GNN to learn exclusively from the subgraphs identified by an explainer. We then measure the model's generalization performance on OOD data splits. This effectively utilizes OOD robustness as a quantitative proxy for the causal validity of the explanation method itself.

\xhdr{Contributions} Our contributions are summarized as follows: (1) We introduce an Explanation-Guided Training framework that incorporates explanatory subgraphs as auxiliary supervision, explicitly encouraging GNNs to focus more on important nodes and improving generalization performance under OOD shifts. (2) We propose \egs, a simple and model-agnostic metric that quantifies the OOD generalization gain from explanation-guided training relative to a vanilla GNN baseline. (3) We provide an extensive empirical study at scale, training and evaluating 11{,}200 models across combinations of datasets (synthetic and real-world), OOD shift settings induced by graph/molecular properties, GNN backbones (GCN, SAGE, GAT, GIN), cross-validation folds, and eight graph explainers, demonstrating both the effectiveness of explanation-guided training and the stability of \egs under changes in backbone and distribution shifts.
\section{Related Work}
\label{sec:related_work}
Our work lies at the intersection of explainability and OOD generalization of GNNs, which we discuss below.

\textbf{Graph Explanations.} Explainability in GNNs has gained significant traction, as it holds the potential for improving trust and interpretability in domains such as drug discovery~\citep{xiong2021graph,li2021graph,huang2023zero}, social network analysis and recommendation systems~\citep{fan2019graph,fan2020graph,gao2023survey,sharma2024survey}. Different explainability methods have been proposed, seeking to understand the internal decision-making processes of GNNs by identifying salient substructures, edges, or node-level features that drive model decisions~\citep{giunchiglia2022towards,agarwal2022probing,agarwal2023evaluating}. Notable examples include GNNExplainer~\cite{ying2019gnnexplainer}, which learns mask-based explanations to isolate predictive subgraphs, and GraphMask~\citep{schlichtkrull2020interpreting}, a post-hoc method for interpreting the predictions of GNNs by learning a mask over messages in GNNs. This body of work collectively aims to make GNN predictions more transparent and help practitioners identify sources of model errors.

\textbf{OOD generalization in graphs.} In parallel, the community has directed considerable effort toward improving GNN generalization, particularly in settings where the training and test distributions significantly differ. 
Consequently, graph invariant learning has emerged as a powerful approach for OOD generalization by isolating invariant features~\cite{li2022learning}.
Studies on Euclidean data have demonstrated that \textbf{leveraging invariant features can provably improve OOD generalization} by relying on the invariance principle: \textit{the existence of features that exhibit stable predictive relationships across different distribution} \cite{2015peters,2019arjovsky,2021kartik}. This principle has been successfully extended to graph data, where the goal is to identify \textit{subgraphs} or \textit{rationales} that capture invariant structural information relevant to the prediction task, leading to improved performance under challenging OOD data.
Notable works in this category include Graph Invariant Learning (GIL)~\cite{li2022learning}, which identifies maximal invariant subgraphs without requiring predefined environment labels and Discovering Invariant Rationale (DIR)~\cite{wu2022invariant_rationales}, which generates interventional distributions to 
identify invariant subgraphs.
A recent work, Cluster Information Transfer (CIT) \cite{xia2024learning}, leverages the invariance principles by transferring nodes across clusters to generate diverse environments, enabling the learning of invariant representations that are robust to structural shifts.


\looseness=-1\xhdr{Evaluation metrics} Several metrics have been proposed for explainability research in GNNs to quantify the effectiveness of an explainer in explaining the underlying model predictions. Please refer to Appendix~\ref{app:metrics} for more details.

\section{OOD Generalization on Graphs}
\label{sec:OOD_gen_graphs}

\subsection{Preliminaries}
\label{sec:prelims}
Throughout our work, an upper-case letter $X$ denotes a random variable, a lower-case letter $x$ indicates a specific instantiation (value) of a random variable, and a calligraphic letter $\mathcal{X}$ indicates a set. Let $G$ represent the random variable of graphs with $g {=} (\mathcal{V}, \mathcal{E}, \mathbf{X})$ representing a single graph with $|\mathcal{V}|$ nodes, set of edges $\mathcal{E}$, and node features $\mathbf{X} = \{\mathbf{x}_0, \mathbf{x}_1, \dots, \mathbf{x}_{n-1}\}$, where $\mathbf{x}_i \in \mathbb{R}^d$, a $d$-dimensional real-valued vector. Next, the adjacency matrix of the graph $g$ is denoted by $\mathbf{A} \in \{0, 1\}^{|\mathcal{V}| \times |\mathcal{V}|}$, where $\mathbf{A}_{uv} = 1$ if $(u,v) \in \mathcal{E}$ ; otherwise $\mathbf{A}_{uv} = 0$. Further, the k-hop subgraph around a given node $u$ is denoted by $\mathcal{S}_u^k = (\mathcal{V}_u^k, \mathcal{E}_u^k, \mathbf{X}_u^k)$.

\xhdr{GNN explanations} In this work, we focus on graph-level explanations, where we explain GNN predictions associated with an entire graph $g$. Formally, the explanation corresponding to graph $g$ that explains the model prediction, \ie $\hat{y}_{g}$, comprises a subset of nodes and edges that influence the prediction $\hat{y}_{g}$. In particular, the generated explanation consists of a discrete node mask $\mathbf{m}_n \in \{0, 1\}^{|\mathcal{V}|}$ and/or an edge mask $\mathbf{M}_e \in \{0, 1\}^{|\mathcal{V}| \times |\mathcal{V}|}$, where an element in $\mathbf{m}_n$ or $\mathbf{M}_e$ takes the value 1 if the corresponding node or edge is identified as important by a GNN explanation method to explain the GNN prediction $\hat{y}_{g}$, and is set to 0 otherwise.

\subsection{Data Generative Model and Invariance}
\begin{figure}[h]
\centering
\vspace{-0.1in}
\begin{tikzpicture}[
    thick,
    every node/.style={draw, circle, minimum size=1cm, font=\sffamily, align=center},
    arrow/.style={-Stealth, thick},
    dashed_arrow/.style={-Stealth, dashed, thick}
]
\node (E) at (-2.5, 2) {$E$};
\node (S) at (0, 2) {$S$};
\node (C) at (1, 0) {$C$};
\node (G) at (-1, 0) {$G$};
\node (Y) at (3, 0) {$Y$};

\draw[arrow] (E) -- (S);
\draw[arrow] (S) -- (G);
\draw[arrow] (C) -- (G);
\draw[arrow] (C) -- (Y);
\draw[dashed_arrow] (S) -- (C);
\draw[dashed_arrow] (C) -- (S);
\end{tikzpicture}
\caption{SCM of data-generating process}
\vspace{-0.24in}
\label{fig:causal_graph}
\end{figure}
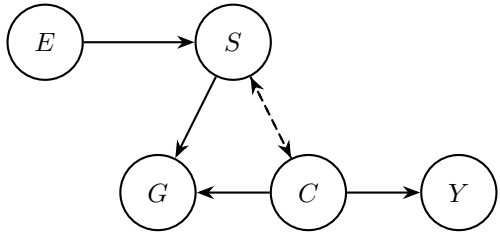

Prior work on invariant causal learning on Euclidean data demonstrates that causally relevant features are invariant across environments and achieve provably OOD generalization \cite{2015peters,2019arjovsky,2021kartik}. For extending this principle to non-Euclidean data, \eg graphs, consider subgraphs as the basic units of causal structure, leveraging their ability to encode localized patterns that remain stable across environments \cite{li2021graph, wu2022invariant_rationales, chen2022learning_causally,li2022learning, xia2024learning}. Following previous works \cite{wu2022invariant_rationales, chen2022learning_causally,li2022learning}, we assume that graphs $G$ are generated from an underlying \textit{Structural Causal Model (SCM)}, which specifies the relationships among a causal subgraph $C$, a spurious subgraph $S$, and the target $Y$ (Fig.~\ref{fig:causal_graph}).
For simplicity, we omit exogenous noise variables, as they are not central to our theoretical argument.

\vspace{0.10in}

Here:\\
(1) The graph $G$ consists of two disjoint components, a causal subgraph $C$ and a spurious subgraph $S$. \\
(2) $E$ represents the environment, \ie any unobserved factors or conditions affecting the data generation process.\\
(3) $C$ is the \textit{causal} subgraph that directly determines $Y$.\\
(4) The dashed arrow $C <---> S$ indicates additional probabilistic dependencies between $C$ and $S$. We consider three types of relationships here: (i) $C$ is independent of $S$, \ie $C \perp S$; (ii) $C$ is a direct cause of $S$, \ie $C \rightarrow S$ (Fully Informative Invariant Features); and (iii) $C$ is an indirect cause of $S$ through $Y$, \ie $C \rightarrow Y \rightarrow S$ (Partially Informative Invariant Features).

This data-generating process accounts for the two most common distribution shifts observed in real-world datasets: \\
(1) Covariate shift, where changes in the distribution of node or edge features ($S$) are induced by variations in the environment ($E$), while the overall graph topology ($G$) remains unchanged. \textit{Example:} In molecular graphs, atomic features may vary across datasets, reflecting differences in experimental conditions or data sources.\\
(2) Structure-level shift: Variations in graph topology, such as changes in edge density or motif distributions, result from shifts in the way the graph ($G$) is generated based on $S$. These structural changes can occur even when $S$ retains similar feature distributions across environments. \textit{Example:} Training graphs might predominantly exhibit tree-like structures, whereas test graphs are grid-like.

An integral assumption of the invariance principle is that the causal relationship $P(Y \mid C)$ \textbf{remains constant or invariant across all environments}, hence, the causal subgraph $C$, retains its predictive power irrespective of spurious correlations in $S$ or variations introduced by $E$ \cite{2015peters,2019arjovsky,2021kartik}. Intuitively, invariant causal subgraphs $C$ are stable predictors across environments because they capture the fundamental mechanisms governing $Y$.

\section{OOD Performance for Quantifying Explanation Quality}
\label{sec:OOD_metric}
\looseness=-1 In Sec.~\ref{sec:OOD_gen_graphs}, we establish that models capturing causal subgraph $C$ (\ie subgraphs invariant across environments) achieve better OOD generalization. Next, we utilize this finding to evaluate the performance of GNN explainers. In particular, we propose a metric that evaluates the quality of explanations generated from GNN explainers by measuring their impact on a model's OOD generalization performance. The underlying idea is to measure the quality of explanations based on how well they correspond to the causal subgraph $C$. To do this, we constrain or guide the optimization of GNNs using a representative explanation obtained from explainers (see Appendix~\ref{app:representative-exp} for information on obtaining representative explanations). In this scenario, models constrained with \textit{causal} explanations are expected to perform better under distributional shifts. Thus, the quality of an explainer can be assessed through the OOD performance of the model it informs.

\textbf{Metric definition.} To evaluate the quality of an explainer, we consider the performance improvement it provides to GNN models under OOD shifts (Fig.~\ref{fig:1}). This involves:

\looseness=-1\textbf{1. Explanation-guided GNNs (EG-GNNs):} The representative explanations from each explainer $\mathtt{E}$ are used to constrain the GNN during training (see Sec.~\ref{sec:aware} for training EG-GNNs), where the constraint guides the GNN to focus on nodes identified as relevant by $\mathtt{E}$.

\textbf{2. Measuring OOD generalization:} The EG-GNN's performance is evaluated on OOD test sets that exhibit distributional shifts from the training data. The relative improvement of the model's performance under these shifts reflects the quality of the explainer $\mathtt{E}$.
    
\textbf{3. Explanation Generalization Score (\egs):} The explainer guiding the GNN to the best OOD generalization performance is considered to have generated the most causal explanations. Mathematically:
    $$
    \text{\egs}(\mathtt{E}) = \frac{\text{Perf}_{\text{OOD}}( \text{GNN}_{\text{EG}}^{\mathtt{E}} ) - \text{Perf}_{\text{OOD}}( \text{GNN}_{\text{base}} )}{\text{Perf}_{\text{OOD}}( \text{GNN}_{\text{base}} )} \times 100\%,
    $$
    where $ \text{Perf}_{\text{OOD}}( \text{GNN}_{\text{EG}}^{\mathtt{E}} )$ is the OOD performance of an EG-GNN using explanations from explainer $\mathtt{E}$, and $ \text{Perf}_{\text{OOD}}( \text{GNN}_{\text{base}} ) $ is the OOD performance of a baseline unconstrained GNN. By computing this metric for different GNN explainers and comparing their respective scores, we identify which explainer generates explanations that correspond the most with the causal subgraphs $C$.

\section{Explanation-Guided GNNs}
\label{sec:ex-gnn}
Here, we first summarize the training of a standard GNN model (Sec.~\ref{sec:standard}) and then describe the formulation of Explanation-Guided GNNs (Sec.~\ref{sec:aware}).
\subsection{Standard GNN Models}
\label{sec:standard}

Formally, GNNs are formulated as message-passing networks comprising three key operators \textsc{Msg}, \textsc{Agg}, and \textsc{Upd}, where the node embeddings are updated iteratively by aggregating information from neighboring nodes. At each layer $l$, node embeddings are updated as:
$$
    \mathbf{h}_{u}^{l+1}{=}\textsc{Upd}_u^{l}(\textsc{Agg}(\textsc{Msg}(\mathbf{h}_{u}^{l},\mathbf{h}_{v}^{l})|v\in\mathcal{N}(u)),\mathbf{h}_{u}^{l})
$$
where $\mathbf{h}_{u}^{l}$ is the embedding of node $u$ at the $l^{\text{th}}$ layer of the GNN, and $\mathcal{N}(u)$ is the set of neighbors of node $u$. The function $\textsc{Upd}_u^{l}$ combines the node's previous embedding with aggregated information from its neighbors $\{\mathbf{h}_j^{l}: j \in \mathcal{N}(u)\}$ using some permutation-invariant aggregation function like summation, mean, or max. 
After several message-passing layers (denoted as $L$ layers), the final embedding for the entire graph, $\mathbf{h}$, is obtained through global pooling on all final node representations $\mathbf{h}_i$. 
These representations are then used to perform graph classification tasks, where we typically use the cross-entropy loss that takes the form: $\mathcal{L} = -\sum_{c=1}^{|\mathcal{Y}|}y_{c} \log(\hat{y}_{c}),$
for a single graph. Here, $y_{c}$ is the true label (one-hot encoded) for class $c$, $\hat{y}_{c}$ is the predicted probability for class $c$, and $|\mathcal{Y}|$ is the number of classes in the graph classification dataset.


\subsection{Explanation-Guided GNNs}
\label{sec:aware}
\looseness=-1 EG-GNNs integrate interpretability into the structure and training of GNNs, enabling models to provide both accurate predictions and insights into their decision-making processes. These architectures leverage explanation methods to guide key operations like message passing and aggregation, filtering out irrelevant connections, and emphasizing important graph substructures. For example, techniques like \textsc{Expass}~\citep{giunchiglia2022towards} and \textsc{DnX}~\citep{pereira2023distill} use node and edge importance scores generated by post-hoc explanations to extract insights from black-box GNNs and distill knowledge into simpler, explainable surrogates, to understand their embeddings and improve trustworthiness.

\looseness=-1 However, a key problem in the above works is that they rely on \textit{post-hoc explanations}, which are generated once the model is trained to achieve a specific predictive performance. Additionally, using post-hoc explanations to improve the underlying GNN could amplify their existing biases and lead to discrepancies in learning the task using the right reasons (that align with domain knowledge)~\citep{ross2017right}. Further, these methods often require complex computations and do not generalize across different datasets~\citep{agarwal2023evaluating}. To address this, for real-world molecular datasets, we consulted domain experts and incorporated annotations that are indicative of specific molecular properties as a regularization function on the model gradients during training, ensuring that the model is inherently interpretable and its predictions align with human domain knowledge. For synthetic datasets, we controlled the data generation process and used ShapeGGen~\citep{agarwal2023evaluating} to create datasets with ground-truth annotations.

\looseness=-1 We work under the hypothesis that models trained using domain-specific knowledge as prior information generalize better to OOD datasets (see Sec.~\ref{sec:OOD_gen_graphs}). Using input-level gradients~\citep{ross2017right,agarwal2022openxai}, we constrain gradients of nodes (annotated by domain knowledge or from the data generating process for synthetic datasets) during the training of the underlying GNN. Now we can either maximize the gradients of relevant nodes identified by the domain experts or minimize the gradients of irrelevant nodes. Since we do not know \textit{a priori} the limit of how large gradients should be, we choose to minimize the gradients (to be zero) of irrelevant nodes.
\begin{equation}
    \begin{split}
    \mathcal{L} &= -\sum_{c=1}^{|\mathcal{Y}|}y_{c} \log(\hat{y}_{c}) + 
    \lambda_1 \sum_{v=1}^{|\mathcal{V}|} \left( \mathbf{a}_{v}^{n} \sum_{c=1}^{|\mathcal{Y}|}\nabla_{v} \log(\hat{y}_{c}) \right)^2 
    \end{split}
    \label{eq:eg-gnn}
\end{equation}
where $\nabla_{v}\log(\hat{y}_{c})$ is the gradient of the predicted probability $\hat{y}_{c}$ of class $c$ for a graph in the dataset w.r.t. the input node. Note that Eqn.~\ref{eq:eg-gnn} can be extended to cases when we have multiple annotations using additional regularization terms for each annotation (see Appendix~\ref{app:datasets} for details).

\xhdr{Domain knowledge annotations} We obtained domain knowledge annotations for specific graph label properties, \ie a subgraph annotation that is indicative of a given graph property (see Appendix~\ref{app:datasets} for more details). Formally, we define a node annotation mask $\mathbf{a}^n \in \{0, 1\}^{|\mathcal{V}|}$, which is a binary mask indicating whether a given node is (not) indicative for predicting a graph property. Note that these annotations can be equivalent to ground truth explanations, \ie they can fully correspond to causal subgraphs for the predictive tasks. 

\section{Experimental Setup}
\label{sec:experimental-setup}
We first describe datasets designed to study the OOD performance of GNNs and then outline the optimization and evaluation setup.

\subsection{Datasets}
We generate two synthetic datasets, \tri and \pen, using ShapeGGen \cite{2022agarwal_evaluating}. ShapeGGen is a synthetic graph generator that generates graphs containing ground-truth motifs using a list of graph-property-related hyperparameters. \tri and \pen contains triangles and pentagons as the ground-truth motifs, respectively. Each dataset contains 750 graph samples, with 50\% being positive samples with ground truth motifs and 50\% being negatives. We obtain a negative graph sample in both datasets by first generating a positive sample, then break all ground-truth motifs within the graph by performing minimal edge removal: for each motif in the graph, we randomly choose one edge in that motif to remove from. This approach ensures the negative samples remain in-distribution compared to the positive samples. We generate graph node features using Laplacian eigenvector positional encoding \cite{dwivedi2023benchmarking}. We additionally use two real-world molecular datasets: \mutag (4,592 samples) and \fluoride (\fc) (1,946 samples). All four datasets in this work provide a range of tasks for assessing the performance of GNNs and their explainers (see Appendix \ref{app:datasets} for more details).

\xhdr{OOD split generation based on graph properties} To evaluate OOD generalization, we generated OOD splits by stratifying datasets according to certain graph properties. For \tri and \pen, we alter three hyperparameters that correspond to graph properties in ShapeGGen to generate OOD splits: probability of edge connection (\textit{ProbConn}), number of motif subgraphs (\textit{NumSubgraphs}), and motif subgraph size (\textit{SubgraphSize}) (see Appendix \ref{app:datasets} for details). For real-world datasets, we generate OOD splits based on molecular properties: weight (\textit{MolWt}), number of rotatable bonds (\textit{NumRB}), fraction of carbon atoms with SP3 hybridization (\textit{FractionCSP3}), and topological polar surface area (\textit{TPSA}). We divide each dataset into 4 OOD splits using near equally-distanced quantiles (herein named as \textit{0to25}, \textit{26to50}, \textit{51to75}, and \textit{76to100}) on each of the corresponding properties mentioned. This results in four OOD splits for each dataset-property pair. This setup enables rigorous evaluation of both model generalization and explanation quality under distribution shifts.

\subsection{Optimization setup}
In our experiments, we use GCN \cite{2016kipf}, GAT \cite{velivckovic2017graph}, SAGE \cite{hamilton2017inductive}, and GIN \cite{xu2018powerful} as the baseline GNN models. During model optimization, given one OOD split, we use 5-fold cross-validation (CV) to optimize model hyperparameters. Each OOD split is divided into train and validation sets at a 4:1 and 8:1 proportion for synthetic and real-world datasets respectively, and we use the validation set to optimize hyperparameters for each CV fold. This results in 5 optimal models for each OOD split. We report each model's performance on the IID validation set, and evaluate its OOD performance on each one of the remaining OOD splits. For example, for training OOD split \textit{0to25}, we obtain 5 IID performance values for \textit{0to25}, and 5 OOD performance values for each of the remaining OOD splits. More optimization and hyperparameter details in Appendix~\ref{app:implementation}.

\subsection{Evaluation Metrics}

\textbf{OOD and IID performance metrics:} We evaluate model performance on both IID and OOD settings using the standard classification metric balanced accuracy. 

\looseness=-1 \textbf{Explanation Generalization Score (\egs):} As discussed in Sec.~\ref{sec:OOD_metric}, \egs quantifies the causal relevance of features identified by explainers and evaluates how well the identified features align with causal subgraphs.
We conduct large-scale experiments, covering two synthetic and two real-world datasets. For each synthetic dataset, we create 12 distinct OOD splits (3 graph properties $\times$ 4 distribution shifts). For real-world datasets, we create 16 distinct OOD splits (4 molecular properties $\times$ 4 distribution shifts). Each split is evaluated using 5-fold CV. We train three types of GNN models: a baseline GNN trained without explanations, an EG-GNN model trained with GTEs (\ie causal subgraphs), and EG-GNN models trained using explanations from 8 different post-hoc explainers. All experiments have been repeated for 4 GNN backbones. \textbf{In total, this results in 11,200 trained models}: (2 datasets $\times$ 16 splits $\times$ 5 folds + 2 datasets $\times$ 12 splits $\times$ 5 folds) $\times$ (1 baseline + 1 GTE + 8 explainers) $\times$ 4 GNNs. For each target OOD split, we report the maximum \egs across all remaining OOD splits as the final \egs. 



\begin{table*}[t]
\centering
\small
\setlength{\tabcolsep}{1pt}
\renewcommand{\arraystretch}{1}
\makebox[\textwidth][c]{
\begin{tabular}{cccccc|cccccc}
\toprule
Property & Synthetic & 0to25 & 26to50 & 51to75 & 76to100 & 
Property & Real-world & 0to25 & 26to50 & 51to75 & 76to100 \\
\midrule

\multirow{2}{*}{NumSubgraph} & \tri & 3.59 & 3.41 & 3.06 & 10.58 &
\multirow{2}{*}{MolWt} & \fc & 0.49 & 0.70 & 3.27 & 5.04 \\
& \pen & 1.42 & 3.35 & 4.55 & 7.76 &
& \mutag & 10.35 & 5.95 & 2.73 & 7.22 \\
\midrule
\multirow{2}{*}{SubgraphSize} & \tri & 8.70 & 3.55 & 4.47 & 6.43 &
\multirow{2}{*}{NumRB} & \fc & 0.33 & 1.00 & 1.72 & 2.40 \\
& \pen & 4.06 & 2.37 & 0.58 & 1.16 &
& \mutag & 3.97 & 4.62 & 3.72 & 0.01 \\
\bottomrule
\end{tabular}
}
\vspace{0.05in}
\caption{\textbf{\egs results using ground-truth explanations. }\egs results using ground-truth explanations across 4 OOD splits (\textit{0to25}, \textit{26to50}, \textit{51to75}, and \textit{76to100}) on synthetic and real-world datasets. For \tri and \pen, we test model generalization performance on \textit{NumSubgraph} and \textit{SubgraphSize} properties; for \fc and \mutag, we test the models on \textit{MolWt} and \textit{NumRB}. We use GAT as the backbone models.}
\label{tab:egsrq1}
\vspace{-0.2in}
\end{table*}


\begin{figure*}[ht]
\begin{center}
\centerline{\includegraphics[width=\textwidth]{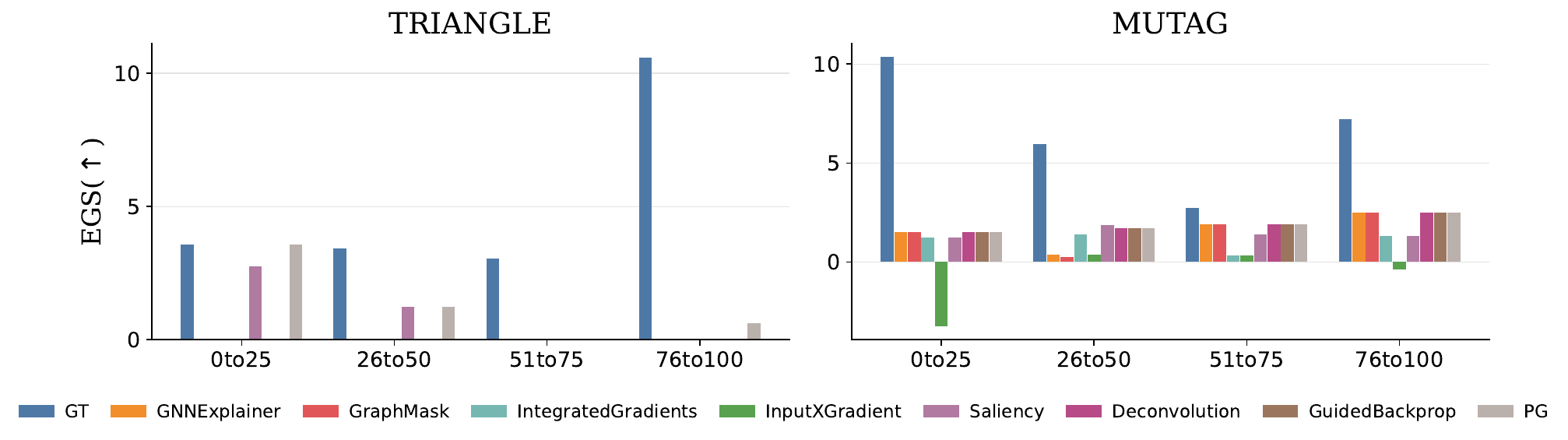}}
\caption{
\textbf{\egs comparison for ground truth and graph explainers.} \egs results using ground-truth explanations and explanations generated from eight graph explainers on \tri (left) and \mutag (right). We evaluate the results over \textit{NumSubgraphs} and \textit{MolWt} properties for \tri and \mutag, respectively. GT indicates Ground Truth Explanations. We use GAT as the backbone models. More results in Appendix~\ref{app:figure}.
}
\label{fig:egsrq2}
\end{center}
\vspace{-0.2in}
\end{figure*} 

\begin{figure*}[ht]
\vskip 0.2in
\begin{center}
\centerline{\includegraphics[width=\textwidth]{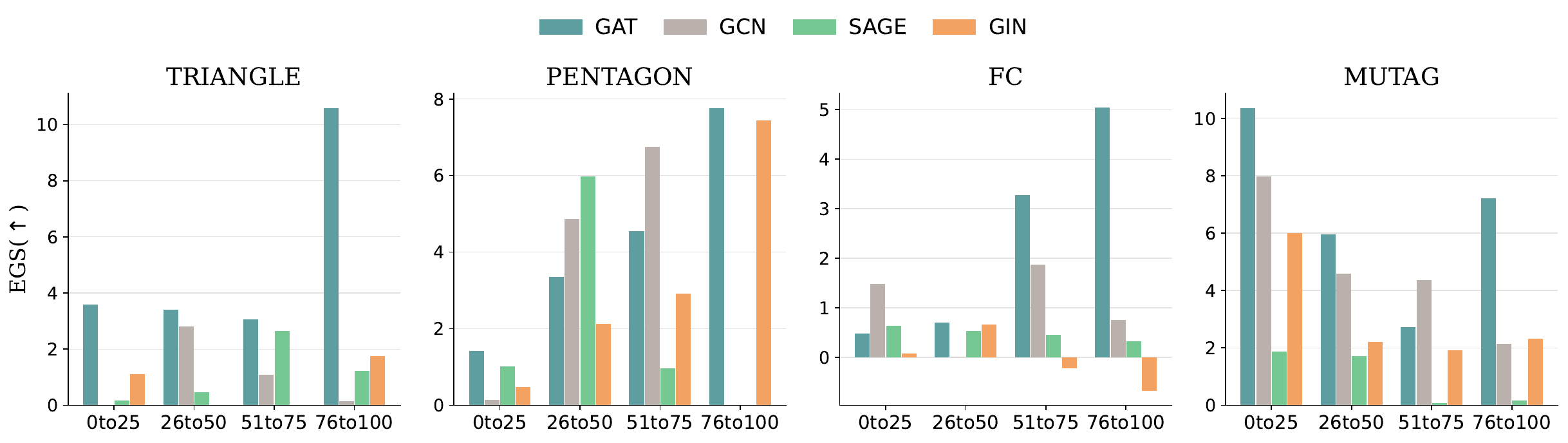}}
\caption{
\textbf{\egs results using ground-truth explanations on all GNN backbones.} ~\looseness=-1 \egs results across four OOD splits on \tri, \pen, \fc, and \mutag using all GNN backbones: GAT, GCN, SAGE, and GIN. We evaluate the results over \textit{NumSubgraphs} and \textit{MolWt} for \tri and \mutag, respectively.
}
\label{fig:egsrq3}
\end{center}
\vspace{-0.3in}
\end{figure*}

\begin{figure*}[ht]
\vskip 0.2in
\begin{center}
\centerline{\includegraphics[width=\textwidth]{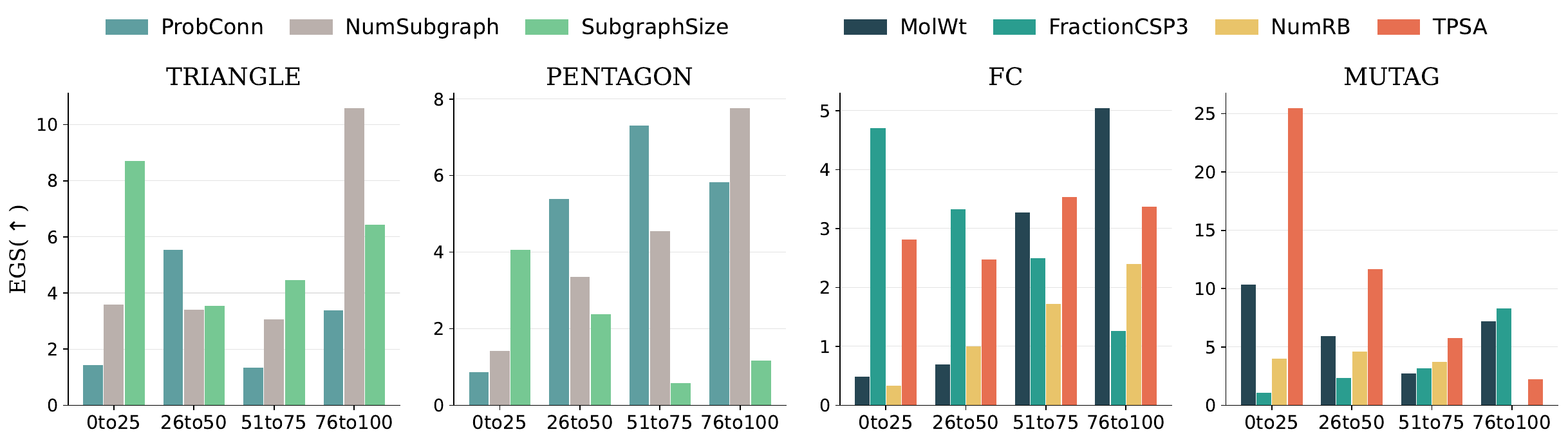}}
\caption{
\textbf{\egs results using ground-truth explanations on all dataset-property pairs.} ~\looseness=-1 \egs results across four OOD splits on \tri, \pen, \fc, and \mutag, using GAT model. For each dataset, we evaluate all of its corresponding graph or molecular properties. 
}
\label{fig:egsrq4}
\end{center}
\vskip -0.2in
\end{figure*}

\section{Results}
\label{sec:results}
\looseness=-1 In this section, we conduct extensive experiments to validate the effectiveness of EG-GNN training under distribution shifts and the utility of \egs metric. We aim to investigate the following research questions: \textbf{RQ1 (Ground-truth explanations):} Under controlled, property-induced distribution shifts, does EG-GNN trained with ground-truth explanations achieve consistently positive \egs across OOD splits compared to the vanilla GNN? \textbf{RQ2 (Explainer comparison):} Across OOD splits and properties, which graph explainers produce explanations that yield the most consistent \egs improvements when used for EG-GNN training? \textbf{RQ3 (GNN Backbone comparison):} Does explanation-guided training produce consistent \egs improvements across different GNN backbones? \textbf{RQ4 (Property Scalability):} How does \egs vary across different graph or molecular properties used to induce OOD splits? 



\xhdr{RQ1) Ground-Truth Explanations Improve OOD Generalization Under Property Shifts} A key premise of this work is that guiding a GNN with causal subgraphs improves robustness under OOD distribution shift. To empirically validate this assumption, we compare the OOD generalization performance of vanilla GNNs and EG-GNNs, where GTEs (\ie causal subgraphs) serve as guidance by minimizing the input gradients of irrelevant nodes. For synthetic datasets, we define irrelevant nodes as those outside the planted motif. For example, on \tri, a node is irrelevant if it does not belong to any triangle subgraph. For real-world datasets, we identify irrelevant nodes through domain expert annotations. We ensure that GTEs correspond to causal explanations by carefully filtering the real-world datasets so that only positive class includes the causal subgraphs (more details in Appendix~\ref{app:datasets}). This ensures that the model remains focused on causal subgraphs during training for both dataset types. 

Table~\ref{tab:egsrq1} records the results of \egs across four OOD splits (\textit{0to25}, \textit{26to50}, \textit{51to75}, and \textit{76to100}), on two synthetic (\tri, \pen) and two real-world (\fc, \mutag) datasets. For synthetic datasets, we evaluate shifts induced by \textit{NumSubgraph} and \textit{SubgraphSize} graph properties. For real-world graphs, we evaluate shifts induced by \textit{MolWt} and \textit{NumRB} molecular properties. Overall, EG-GNNs trained with ground-truth explanations achieve positive \egs on every split, property, and dataset, indicating that explanation guidance consistently improves OOD generalization performance relative to the vanilla GNN. For \tri, we observe the highest performance gains, with \egs value of 10.58 on \textit{76to100} split of \textit{NumSubgraph}. For \mutag, the highest \egs value is at \textit{0to25} split of \textit{MolWt}. This reinforces that explanations leading to better OOD performance (\ie have higher \egs) may highlight the most functionally relevant substructures, making them valuable for building hybrid models that integrate domain knowledge into predictions. We further verify that the ground-truth masks used for EG-GNN training remain aligned with the model’s decision evidence by reporting faithfulness scores in Figure~\ref{fig:faithfulness}. Across all four datasets, faithfulness is consistently high (near-saturated on \fc and \mutag), indicating that EG-GNN primarily relies on the annotated causal subgraphs when making predictions.

\xhdr{RQ2) \egs comparison of graph explainers} We investigate \egs results of EG-GNNs on OOD splits using eight different graph explainers. We form the final explanation mask for each explainer by finding the most common connected subgraph based on an explainer's attribution scores. We then perform matching within each positive sample for every occurrence of this representative subgraph, and mark all matching nodes as important. For synthetic datasets, the minimum number of nodes in a connected subgraph is set to 3, and 2 for real-world datasets (more details about explainer explanation generation process in Appendix~\ref{app:datasets}). The final explanations are then used to train EG-GNNs in place of ground-truth explanations to guide the learning process.

Figure~\ref{fig:egsrq2} reports \egs on two dataset–property pairs: \tri–\textit{NumSubgraphs} and \mutag–\textit{MolWt}. In both cases, ground-truth explanations achieve the largest OOD gains, serving as an upper bound on explanation-guided training. On \mutag, most explainers yield positive \egs across splits, but remain substantially below GT, with the largest gaps at \textit{0to25}, \textit{26to50}, and \textit{76to100}. This suggests that explainer-generated masks recover part of the causal signal but introduce noise that limits the achievable improvement. Notably, Input$\times$Gradient receives negative results under two OOD shifts (\textit{0to25}, \textit{76to100}), indicating that inaccurate masks may misguide training. 

\looseness=-1 For \tri, we observe that a large portion of explainers have a \egs value of zero, which mainly reflects our pattern-extraction and matching pipeline. With a minimum connected-subgraph size set to 3, the extracted representative subgraph often collapses to a generic 3-node path. In synthetic graphs, such a path is highly non-specific and matches broadly, regardless of whether a node participates in a causal motif (triangle or pentagon). In contrast, on molecular graphs, even small connected patterns (2-node or 3-node fragments) can carry chemically meaningful identity through atom and bond types (more details in Appendix~\ref{app:representative-exp}). In such scenarios, all nodes in the graph are marked as important, leaving EG-GNN with zero nodes labeled as unimportant to minimize. This is equivalent to vanilla GNN training. Despite this, Saliency and PG explainers show non-zero gains for \textit{0to25}, \textit{26to50}, and \textit{76to100}. This implies that when an explainer happens to induce a more meaningful motif, EG-GNN training can recover measurable OOD improvements.

\xhdr{Ablation Study}
We explore the scalability of EG-GNN training using ground-truth explanations as guidance. Specifically, we explore whether \egs remains a stable indicator of EG-GNN when scaling along two axes: changing the GNN backbones (\textbf{RQ3}), and changing the property used to induce the OOD splits (\textbf{RQ4}). 

Figure~\ref{fig:egsrq3} compares \egs across four GNN backbones over OOD splits for each dataset. For synthetic datasets, the evaluated property is \textit{NumSubgraphs}, while for real-world datasets, it is \textit{MolWt}. We observe non-negative \egs values across three GNN backbones on all datasets, indicating that explanation guidance transfers across architectures in these settings. At the same time, GAT shows the strongest and most consistent improvements, while other backbones exhibit smaller or more variable gains. This indicates that despite transferable, the same supervision signal can translate into different degrees of OOD benefit depending on backbone inductive bias and capacity. 

For \fc, GIN yields negative \egs at \textit{51to75} and \textit{76to100}, suggesting that explanation guidance can occasionally be harmful when the provided masks are less precise or less complete. This scenario occurs more often for real-world data than synthetic because the ground-truth explanations are based on expert annotations and dataset curation rather than a fully controlled generative process. Hence, the annotated substructures may not capture all predictive evidence in every split and there may be some spurious correlation in the dataset that the underlying GNN can capture to get high predictive performance~\citep{faber2021comparing}. In addition, the difference in performance between GAT and GIN suggests that the GNN architecture might be fundamentally at odds with the explanation signal, \ie if an explanation mask highlights specific nodes, but the GNN backbone (like GIN) relies heavily on global graph isomorphisms or specific structural aggregations that the mask disrupts, the ``guidance'' acts as noise.

Fig.~\ref{fig:egsrq4} evaluates \egs across multiple dataset--property pairs using a fixed backbone. For synthetic datasets, we consider three graph properties (ProbConnection, NumSubgraphs, and SubgraphSize), and for real-world datasets we consider four molecular properties (MolWt, FractionCSP3, NumRB, and TPSA). Overall, \egs remains non-negative across these dataset--property pairs, showing that EG-GNN with ground-truth guidance consistently improves OOD generalization across a broad range of shift generators. We also observe that the magnitude of \egs varies by property and split (\eg larger gains under certain shifts on \mutag), which suggests that \egs can differentiate when explanation guidance is most effective under different OOD constructions.

\section{Conclusion}
\label{sec:discussion}
The proposed framework for evaluating OOD generalization through explanation-guided GNNs (EG-GNN) provides a novel perspective on assessing the quality of GNN explainers. Unlike traditional metrics which often rely on static comparisons, this approach directly evaluates the functional utility of explanations in improving model performance under distributional shifts. By focusing on the relative improvement achieved by EG-GNNs, the framework offers a robust and scalable method to identify explainers that produce more generalizable and domain-relevant explanations.
Importantly, \egs decouples the evaluation of explainers from the absolute performance of the baseline GNN, addressing a critical limitation in current practices where poor baseline performance can obscure the value of an explainer. 
The focus on OOD performance is particularly relevant in real-world applications, such as drug discovery or material design, where models are frequently exposed to unseen or shifted data distributions. 

\section*{Acknowledgements}
The authors would thank Changpeng Lu and Guadalupe Gonzalez for their extensive, helpful discussions and feedback throughout idea formation and experiment validations. C.A. and Aikyam Lab is supported, in part, by grants from Capital One, LaCross Institute for Ethical AI in Business, the UVA Environmental Institute, OpenAI Researcher Program, Thinking Machine’s Tinker Research Grant, and Cohere. The views expressed are those of the authors and do not reflect the official policy or the position of the funding agencies.

\bibliographystyle{unsrtnat}
\bibliography{reference}

\appendix
\section{Appendix}
\subsection{Datasets}
\label{app:datasets}

Synthetic Datasets: We construct the \tri and the \pen dataset using ShapeGGen \cite{2022agarwal_evaluating}, a synthetic graph generator that generates graphs containing ground-truth motifs. To induce controlled distribution shifts, we create OOD splits by varying three ShapeGGen hyperparameters that directly correspond to graph-level properties: probability of edge connection (ProbConn), number of motif subgraphs (NumSubgraph), and motif subgraph size (SubgraphSize). ProbConn controls the probability that two motifs are connected. A high ProbConn value indicates a motif will have high probability connecting to another motif. NumSubgraph controls how many motifs are inserted into each graph. SubgraphSize controls the size of each motif by attaching additional nodes to the base motif. For example, when SubgraphSize is set to 5 for \tri, each triangle motif is augmented with two extra nodes connected to the triangle, increasing the motif’s local neighborhood while preserving the underlying triangle structure. 

When generating each graph, we sample the hyperparameter of the target OOD property shift from a pre-defined range, while keeping the other two hyperparameters fixed to their default settings. ProbConn is sampled from a continuous range, whereas NumSubgraph and SubgraphSize are sampled from discrete integer ranges. The specific sampling ranges used to define the four OOD splits for \tri and \pen are summarized in Table~\ref{tab:syn_hyper}.

\begin{table}[H]
\centering
\small
\setlength{\tabcolsep}{4pt}
\begin{tabular}{lcccccccc}
\toprule
\multirow{2}{*}{} 
& \multicolumn{4}{c}{\tri} 
& \multicolumn{4}{c}{\pen} \\
\cmidrule(lr){2-5}\cmidrule(lr){6-9}
& 0to25 & 26to50 & 51to75 & 76to100
& 0to25 & 26to50 & 51to75 & 76to100 \\
\midrule

\multirow{1}{*}{ProbConn}     & [0.05, 0.2] & [0.3, 0.4] & [0.6, 0.7] & [0.9, 1.0] & [0.05, 0.2] & [0.3, 0.4] & [0.6, 0.7] & [0.9, 1.0] \\
\multirow{1}{*}{NumSubgraph}  & (2, 3) & (6, 7) & (9, 10) & (12, 13) & (2, 3) & (5, 6) & (8, 9) & (11, 12) \\
\multirow{1}{*}{SubgraphSize} & (3, 4) & (6, 7) & (9, 10) & (12, 13) & (6, 7) & (9, 10) & (12, 13) & (15, 16) \\

\bottomrule
\end{tabular}

\caption{\textbf{Hyperparameter ranges for constructing OOD splits on synthetic datasets.} 
ShapeGGen hyperparameter ranges used to define the four OOD splits (\textit{0to25}, \textit{26to50}, \textit{51to75}, \textit{76to100}) for \tri and \pen. Each entry indicates the sampling range of the corresponding hyperparameter when generating graphs within that split (brackets denote continuous ranges; parentheses denote integer ranges).}
\label{tab:syn_hyper}
\end{table}

\fc: We obtained the \fluoride dataset from \citet{2022agarwal_evaluating}. The \fc dataset contains 8,671 molecular graphs labeled into two classes where positive samples more frequently contain a fluoride (\verb|[F]|) and a carbonyl (\verb|[C](=O)|) functional group. The ground-truth explanations, or causal graph, hence consist of combinations of fluoride atoms and carbonyl functional groups. From this dataset, we excluded both, molecules labeled as positive but lacking the causal subgraph, and molecules labeled as negative but containing the causal subgraph. Additionally, we down-sampled negative samples to achieve a positive ratio of 30\%, as lower ratios resulting in extremely unbalanced OOD splits. This resulted in 1,946 samples with 584 positive examples and 1,362 negative examples.

\mutag: We obtained the \mutag dataset from \citet{hansen2009benchmark}, a benchmark dataset merging six data sources, for the prediction of Ames mutagenicity. Ames mutagenicity refers to the ability of a chemical compound to induce mutations in the DNA of bacteria, as assessed by the Ames test. We extracted a list of substructures predominantly found in the positive class from \citet{2005kazius}. We filtered out both, molecules labeled as positive but lacking the causal subgraph, and molecules labeled as negative but containing the causal subgraph. This resulted in 4,592 samples with 2,252 positive examples and 2,340 negative examples. Positive samples contain either of the following causal subgraphs: (1) 5-atom aromatic nitro: \verb|a1aaa(aa1)~[NX3](-,=[OX1])-,=[OX1]|; (2) 6-atom aromatic nitro: \verb|a1(~[NX3](-,=[OX1])-,=[OX1])aaaa1|; (3) three-membered heterocycle: \verb|C1C[NH,O,S]1|; (4) nitroso: \verb|N=O|; (5) unsubstituted heteroatom-bonded heteroatom: \verb|[N,O]-[NH2,OH]|; (6) azo-type: \verb|N=N|; (7) aliphatic alide: \verb|[Cl,Br,I][C;!\$(C=*)]|; (8) 6-6-6 polycyclic aromatic system: \verb|a2aa1a(aaaa1)a3a2aaaa3|; (9) 5-6-6 polycyclic aromatic system: \verb|a2a1a(aaa1)aa3a2aaaa3|; (10) 6-5-6 polycyclic aromatic system: \verb|a1a3a(a2a1aaaa2)aaaa3|.


\subsection{Implementation Details}
\label{app:implementation}

For baseline GNN models we optimize the dimension of hidden layers (16, 32, 64, 128), the learning rates (5e-4, 0.001, 0.005, 0.01), and batch size (4, 8, 16). For EG-GNN models we additionally optimize the explanation loss ($\lambda_1$) sampled within the range [1e-4, 1.0] while keeping other hyperparameters unchanged. We implemented our model in PyTorch 2.7.0~\citep{paszke2019pytorch} and torch-geometric 2.6.1~\citep{Fey/Lenssen/2019}. We provide a list of optimized hyperparameter choices in Table~\ref{tab:hparams}.

\begin{table}[t]
\centering
\small
\setlength{\tabcolsep}{5pt}
\renewcommand{\arraystretch}{0.95}
\begin{tabular}{llccccc}
\toprule
Dataset & Backbone & Batch size & LR & Hidden dim & Epochs & $\lambda_1$ (EG-GNN) \\
\midrule

\multirow{4}{*}{\tri}
& GCN  & 8  & 0.001 & 64  & \multirow{4}{*}{50} & \multirow{4}{*}{0.3} \\
& SAGE & 4  & 0.001 & 128 &  &  \\
& GAT  & 8  & 0.005 & 64  &  &  \\
& GIN  & 8  & 0.001 & 64  &  &  \\
\midrule

\multirow{4}{*}{\pen}
& GCN  & 16 & 0.001  & 128 & \multirow{4}{*}{50} & \multirow{4}{*}{0.2} \\
& SAGE & 16 & 0.0005 & 128 &  &  \\
& GAT  & 4  & 0.0005 & 64  &  &  \\
& GIN  & 16 & 0.0005 & 128 &  &  \\
\midrule

\multirow{4}{*}{\fc}
& GCN  & 16 & 0.01  & 64  & \multirow{4}{*}{30} & \multirow{4}{*}{0.1} \\
& SAGE & 16 & 0.001 & 64  &  &  \\
& GAT  & 16 & 0.005 & 128 &  &  \\
& GIN  & 16 & 0.005 & 128 &  &  \\
\midrule

\multirow{4}{*}{\mutag}
& GCN  & 8  & 0.005 & 64  & \multirow{4}{*}{30} & \multirow{4}{*}{0.3} \\
& SAGE & 8  & 0.005 & 64  &  &  \\
& GAT  & 16 & 0.005 & 128 &  &  \\
& GIN  & 16 & 0.005 & 128 &  &  \\
\bottomrule
\end{tabular}

\caption{\textbf{Training hyperparameters for vanilla GNNs and EG-GNNs.} We report the batch size, learning rate (LR), hidden dimension, and number of training epochs for each dataset--backbone pair. For EG-GNN, we additionally report the explanation-loss weight $\lambda_1$, which is fixed across backbones within each dataset.}
\label{tab:hparams}
\end{table}


\subsection{Explanation Frameworks}
\label{app:explanation}
Our framework incorporates eight distinct explanation methods, each offering a unique approach to interpreting GNNs. These methods can be categorized based on their computational mechanism (gradient-based or perturbation-based) and their explanation type (post-hoc, factual, or counterfactual).

\subsubsection{Graph-Specific Explainers}
\label{app:explainers}
\textbf{GNNExplainer}~\citep{ying2019gnnexplainer} is a post-hoc, perturbation-based method that identifies the most important subgraph structures and node features for a given prediction. It learns a soft mask over edges and features to maximize mutual information between the model’s prediction and the selected subgraph. This method is particularly effective for discovering recurring motifs and structural dependencies, making it valuable for tasks such as molecular property prediction and community detection in social networks.

\textbf{GraphMask}~\citep{schlichtkrull2020interpreting} is a post-hoc, perturbation-based method that learns a sparse mask over the messages passed between nodes. Unlike GNNExplainer, which identifies subgraphs, GraphMask operates at the message-passing level, determining which node interactions are most influential for a model’s decision. This makes it particularly useful for analyzing information flow in dynamic and multi-relational graphs.

\textbf{PGExplainer}~\citep{luo2020parameterized} (Parameterized Graph Explainer) advances perturbation-based explanation by learning a parameterized model to generate explanations collectively across multiple instances. PGExplainer trains a neural network to predict the probability of edge importance based on node embeddings. This parameterized approach allows it to provide explanations with a global view of the GNN model and also improves efficiency, particularly in inductive settings where explanations must be generated for new, unseen graphs without retraining.

\subsubsection{Gradient-Based Attribution Methods}

We utilize five gradient-based explainers implemented via the Captum framework~\citep{kokhlikyan2020captum}:

\textbf{Integrated Gradients}~\citep{sundararajan2017axiomatic} computes attributions by integrating gradients along a linear interpolation between a baseline input and the actual input. It satisfies key theoretical properties such as sensitivity and implementation invariance, making it particularly effective for capturing cumulative feature importance across the entire input space. The choice of baseline allows for limited counterfactual reasoning.

\textbf{Saliency Maps}~\citep{simonyan2013deep} represent a first-order gradient-based method that estimates node importance by computing the absolute gradient of the model's output with respect to input features. While simple, it highlights nodes with the most immediate impact on the output prediction and serves as a useful baseline for comparison with more structured methods.

\textbf{Input $\times$ Gradient}~\citep{shrikumar2016black} refines raw gradient-based attribution by weighting gradients by their corresponding input values. This helps distinguish between nodes that are inherently important (high magnitude) and those that are particularly sensitive to model predictions.

\textbf{Deconvolution}~\citep{zeiler2013visualizing} modifies the standard backpropagation process by restricting gradient propagation through ReLU activations, ensuring that only positive activations are passed during explanation. This results in clearer attribution maps, particularly for models with complex activation functions.


\textbf{Guided Backpropagation}~\citep{springenberg2014striving} enhances traditional backpropagation by modifying gradient flow such that only positive contributions propagate through ReLU activations. This results in sharper, more interpretable visualizations of node importance, making it especially effective for identifying hierarchical feature importance in structured graphs.

Each of these methods provides a distinct perspective on GNN decision-making, allowing for a robust interpretability analysis. By integrating perturbation-based, gradient-based, and game-theoretic approaches, we ensure a comprehensive understanding of how structural and feature-based properties influence model predictions.

\subsection{Representative Explanation for GNN Explainers}
\label{app:representative-exp}

To identify the most representative causal subgraphs highlighted by different graph explainers, we extract and analyze frequently occurring subgraphs from explanation masks. We obtain explanations from a set of explainers designed to interpret baseline GNNs trained in an IID setting, ensuring that explanations reflect model behavior under standard training conditions. This process is repeated for each OOD split (\ie \textit{0to25}, \textit{26to50}, \textit{51to75}, \textit{76to100}) to capture how explanations vary across different data distributions.

For every graph, we extract the most relevant substructures based on an explainer's attribution scores. Specifically, for each graph $g = (\mathcal{V}, \mathcal{E})$ in our dataset, we generate explanations following a systematic protocol:

\begin{enumerate}
    \item For each explainer $\mathtt{E}$, we compute node importance scores $s_v \in \mathbb{R}$ for all $v \in \mathcal{V}$.
    \item For gradient-based explainers operating on node features $\mathbf{X} \in \mathbb{R}^{|\mathcal{V}| \times d}$, we aggregate feature-level importance scores across the feature dimension:
    \begin{equation}
        s_v = \frac{1}{d}\sum_{i=1}^d s_{v,i}
    \end{equation}
    where $s_{v,i}$ represents the importance score for feature $i$ of node $v$. Note that for PGExplainer \cite{luo2020parameterized}, the importance scores are on edges rather than nodes. To make the explanation compatible, we convert edge attributions into node attributions by treating a node as important if it is incident to at least one high-importance edge.
\end{enumerate}

To convert continuous importance scores into binary node masks, we employ a threshold-based approach. Specifically, we retain the top-$k$ most important nodes, where $k = \lceil 0.3 \times |\mathcal{V}| \rceil$, creating a binary node mask $\mathbf{m}_n \in \{0,1\}^{|\mathcal{V}|}$. This approach ensures that approximately 30\% of nodes are identified as important for each explanation, providing a consistent basis for comparison across different graphs and explainers.

Explanations for each OOD split are generated across cross-validation folds. To extract a stable “representative” pattern for each explainer, we collect the connected subgraphs induced by the binary masks and select the most frequently occurring connected substructure across folds, subject to a minimum size constraint. For synthetic datasets (\tri and \pen), we require at least 3 nodes. We use 3 (rather than 2) because synthetic nodes do not carry semantic labels, and a 2-node fragment is simply an edge that appears in nearly all graphs, which would lead to non-informative patterns and overly dense masks. With a minimum of 3 nodes, the representative subgraph can capture motif-specific structure (for example, a triangle fragment) and is less likely to degenerate into a trivial pattern. For real-world molecular datasets (\fc and \mutag), we allow 2-node subgraphs because even small fragments can be meaningful due to atom and bond types; a 2-node pattern corresponds to a specific bond between specific atom types and is therefore less likely to match indiscriminately.

Given the representative subgraph, we then construct the final explanation mask for each graph by marking all occurrences of that subgraph as important. For synthetic graphs, we search for all subgraph isomorphism matches and set the nodes in every matched occurrence to 1. This procedure can yield an all-ones mask in the degenerate case where the representative pattern is a generic 3-node path, since such paths can appear widely throughout the graph. For molecular graphs, we use the associated SMILES representation to find all occurrences of the representative fragment (as a chemically grounded substructure match), and mark the matched atoms as important in the corresponding molecular graph. The resulting binary masks are used as the explainer-derived explanations for each combination of dataset, property, OOD split, and explainer.

\subsection{Evaluation Metrics}
\label{app:metrics}

\begin{itemize}
    \item \textsc{Fidelity:} Measures how well the explanation (\eg important nodes/edges) correlates with the GNN's performance. High fidelity implies that masking out the explained parts significantly changes the prediction~\cite{yuan2022explainability}. 
    We use $\text{Fidelity}_+$ to study the prediction change by removing important nodes/edges/node features, where:
    \textit{Fidelity-Plus ($\text{Fid}_+$):} Measures how well the important subgraph preserves the original prediction:
    \begin{equation}
        \text{Fid}_+ = 1 - \mathbb{E}_{g \in \mathcal{G}}[\mathbbm{1}(f(g_{\mathbf{m}}) = f(g))].
    \end{equation}
    
    \textit{Fidelity-Minus ($\text{Fid}_-$):} Quantifies the impact of removing important nodes:
    \begin{equation}
        \text{Fid}_- = 1 - \mathbb{E}_{G \in \mathcal{G}}[\mathbbm{1}(f(G_{\neg\mathbf{m}}) = f(G))]
    \end{equation}

    \item 
    \textsc{Characterization Score ($\mathcal{C}$) :}~\citet{amara2022graphframex} proposes characterization score as an analogous to combining precision and recall in the Micro-F1 metric, where the metric is calculated as the weighted harmonic mean of Fidelity+ and (1 - Fidelity-).
    \begin{equation}
        \mathcal{C} = \frac{w_+ + w_-}{\frac{w_+}{\text{Fid}_+} + \frac{w_-}{1-\text{Fid}_-}}.
    \end{equation}
    
    \item \textsc{Unfaithfulness ($\mathcal{U}$):} It extends existing faithfulness metrics to quantify how faithful explanations are to an underlying GNN predictor by obtaining the prediction probability vector using the GNN and using the explanation (generate a masked subgraph by only keeping the original values of the top-k features identified by an explanation)~\cite{agarwal2023evaluating}. Finally, the unfaithfulness metric is calculated using the normalized Kullback-Leibler (KL) divergence score that quantifies the distance between two probability vectors.
    \begin{equation}
        \mathcal{U} = 1 - \exp(-D_{\text{KL}}(P_{\text{orig}} \| P_{\text{expl}})).
    \end{equation}
\end{itemize}

\subsection{Figures}
\label{app:figure}

\begin{figure*}[h]
\vskip 0.2in
\begin{center}
\centerline{\includegraphics[width=\textwidth]{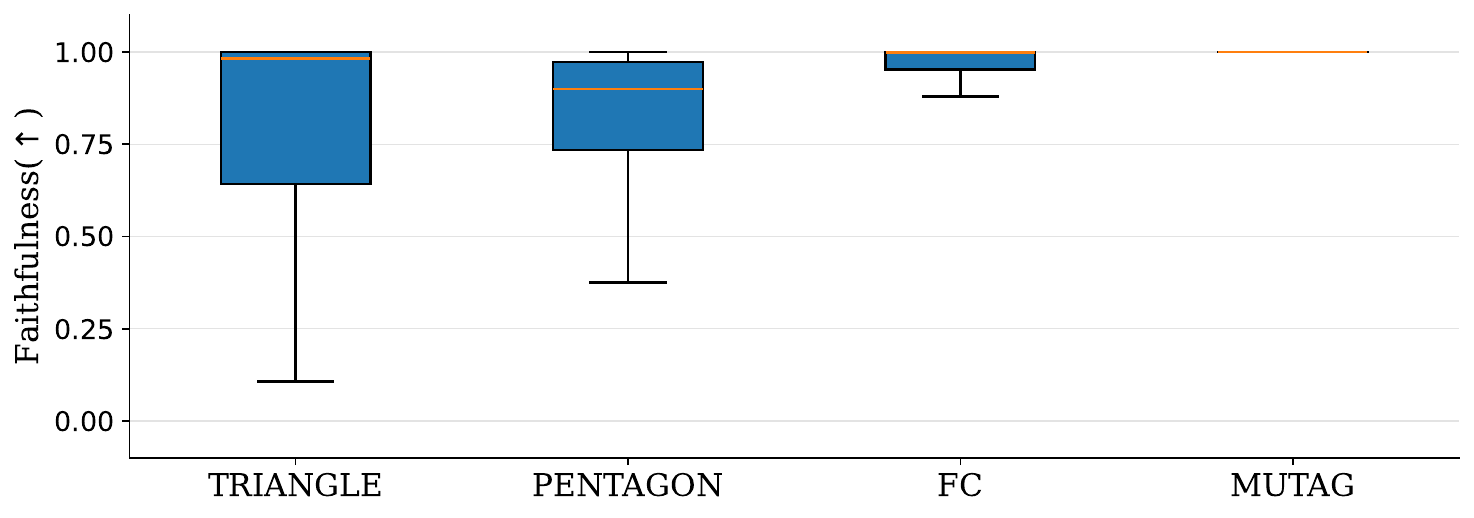}}
\caption{\textbf{Faithfulness of ground-truth explanations for EG-GNNs across four datasets.}
Distribution of faithfulness scores computed using ground-truth explanations for EG-GNNs on \tri, \pen, \fc, and \mutag. Each box summarizes faithfulness over all test samples for all OOD splits, properties, and GNN backbones. The central line in each box denotes the median; boxes indicate the interquartile range (IQR); whiskers extend to 1.5$\times$IQR; and outlier points are omitted for readability.}
\label{fig:faithfulness}
\end{center}
\vskip -0.2in
\end{figure*} 

\begin{figure}[H]
\vskip 0.2in
\begin{center}
\centerline{\includegraphics[width=\textwidth]{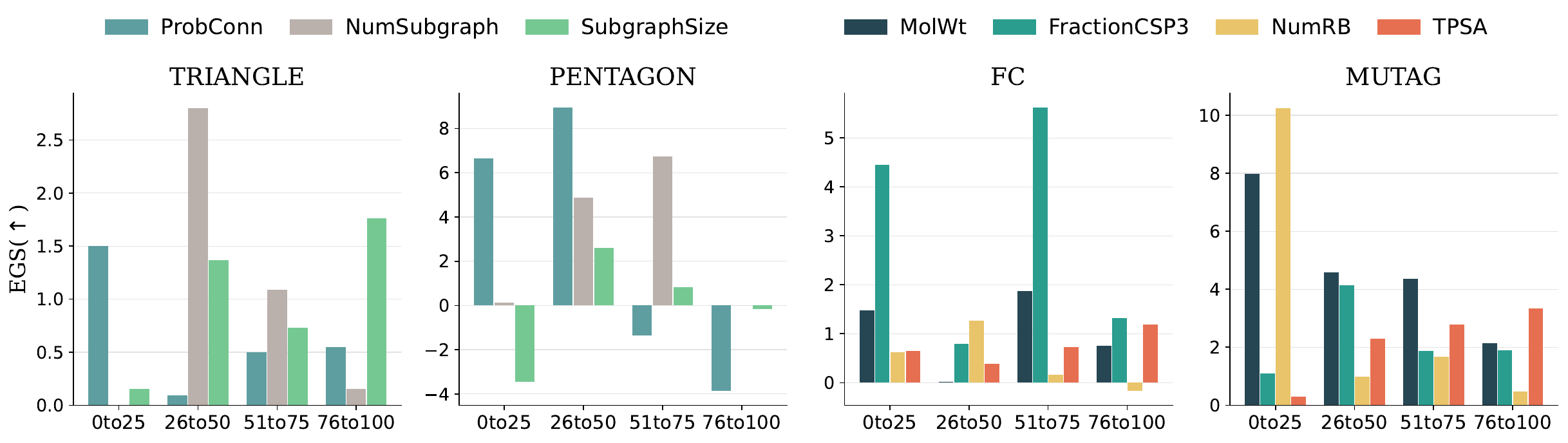}}
\caption{
\textbf{\egs results using ground-truth explanations on all dataset-property pairs using GCN.} ~\looseness=-1 \egs results using ground-truth explanations across four OOD splits on \tri, \pen, \fc, and \mutag, using GCN model. For each dataset, we evaluate all of its corresponding graph or molecular properties. 
}
\end{center}
\vskip -0.2in
\end{figure}

\begin{figure}[H]
\vskip 0.2in
\begin{center}
\centerline{\includegraphics[width=\textwidth]{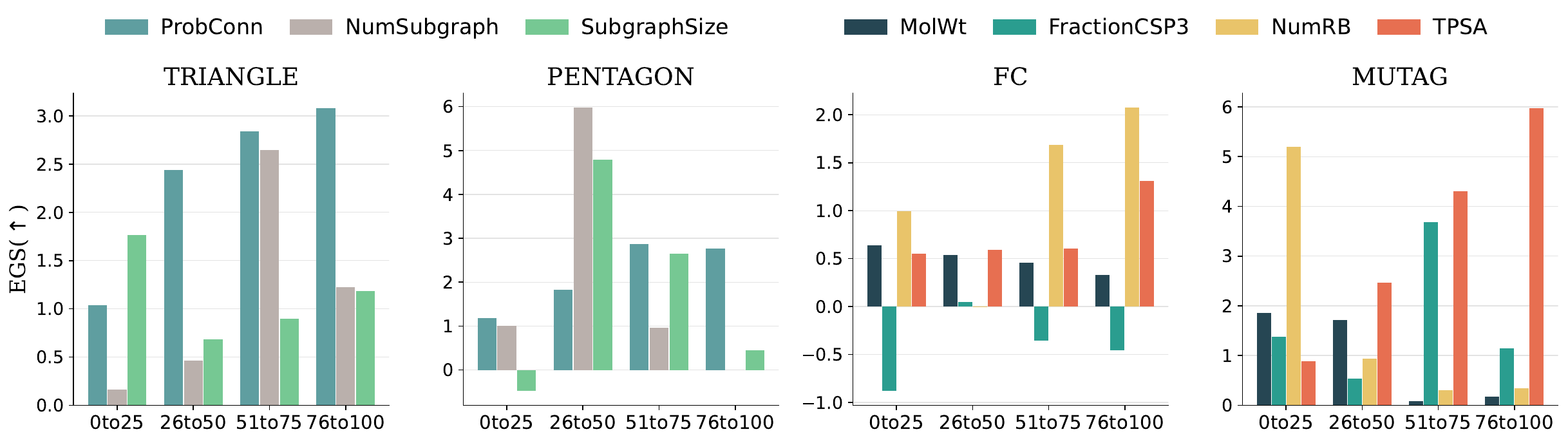}}

\caption{
\textbf{\egs results using ground-truth explanations on all dataset-property pairs using SAGE.} ~\looseness=-1 \egs results using ground-truth explanations across four OOD splits on \tri, \pen, \fc, and \mutag, using SAGE model. For each dataset, we evaluate all of its corresponding graph or molecular properties. 
}
\end{center}
\vskip -0.2in
\end{figure} 

\begin{figure}[H]
\vskip 0.2in
\begin{center}
\centerline{\includegraphics[width=\textwidth]{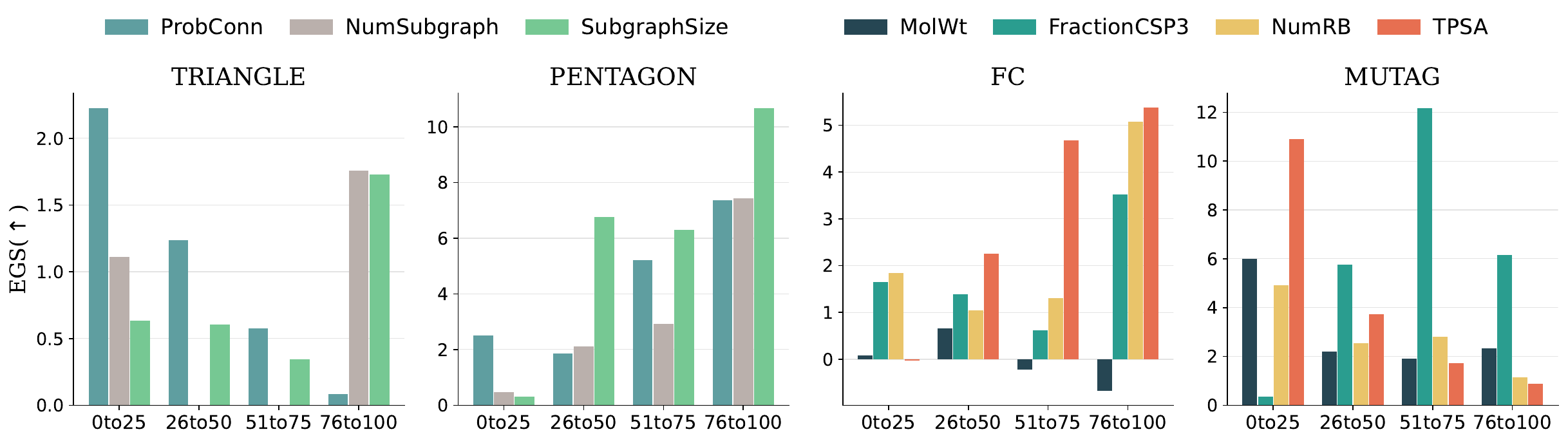}}

\caption{
\textbf{\egs results using ground-truth explanations on all dataset-property pairs using GIN.} ~\looseness=-1 \egs results using ground-truth explanations across four OOD splits on \tri, \pen, \fc, and \mutag, using GIN model. For each dataset, we evaluate all of its corresponding graph or molecular properties. 
}
\end{center}
\vskip -0.2in
\end{figure}

\begin{figure*}[h]
\vskip 0.2in
\begin{center}
\centerline{\includegraphics[width=\textwidth]{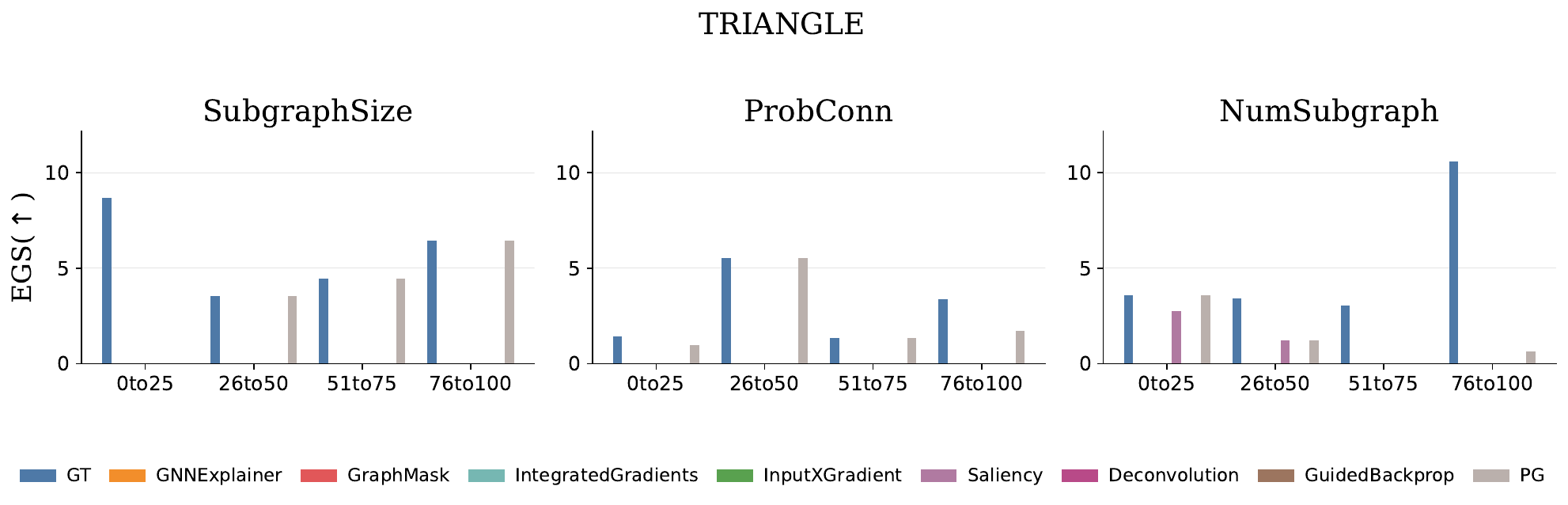}}

\caption{
\textbf{\egs comparison for ground truth and graph explainers using GAT.} ~\looseness=-1 \egs results of explanation-guided training using ground-truth explanations and explanations generated from eight graph explainers on \tri over all graph properties.  
}
\end{center}
\vskip -0.2in
\end{figure*}

\begin{figure*}[h]
\vskip 0.2in
\begin{center}
\centerline{\includegraphics[width=\textwidth]{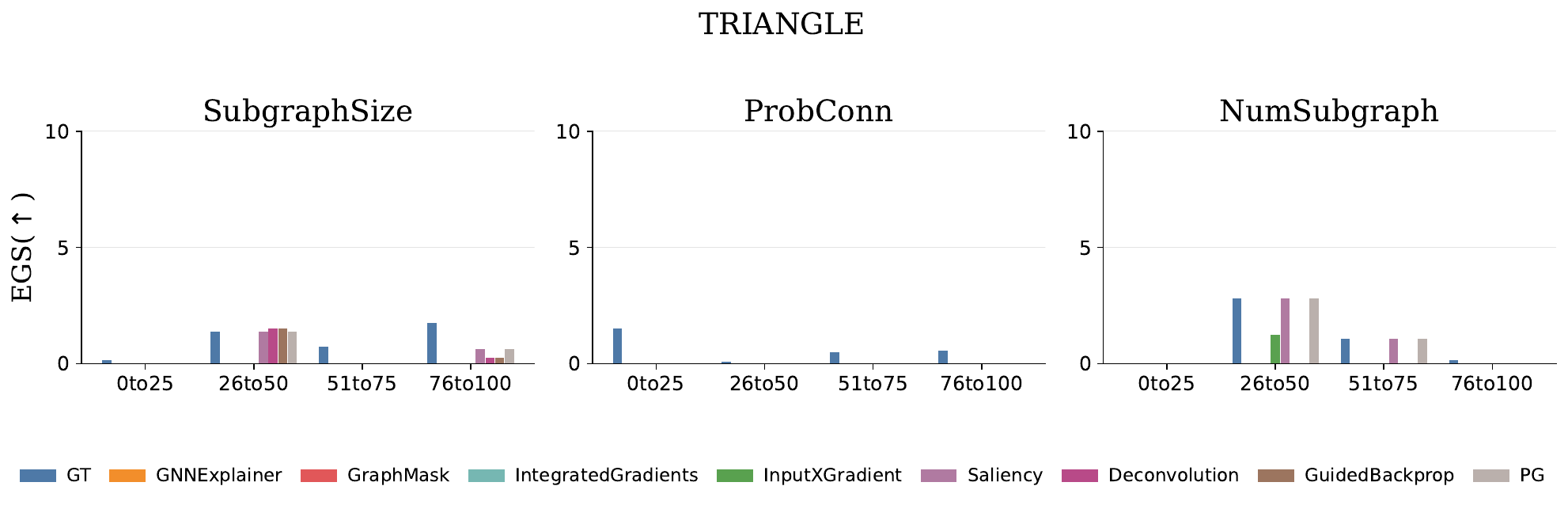}}

\caption{
\textbf{\egs comparison for ground truth and graph explainers using GAT.}  ~\looseness=-1 \egs results of explanation-guided training using ground-truth explanations and explanations generated from eight graph explainers on \tri over all graph properties.  
}
\end{center}
\vskip -0.2in
\end{figure*}

\begin{figure*}[h]
\vskip 0.2in
\begin{center}
\centerline{\includegraphics[width=\textwidth]{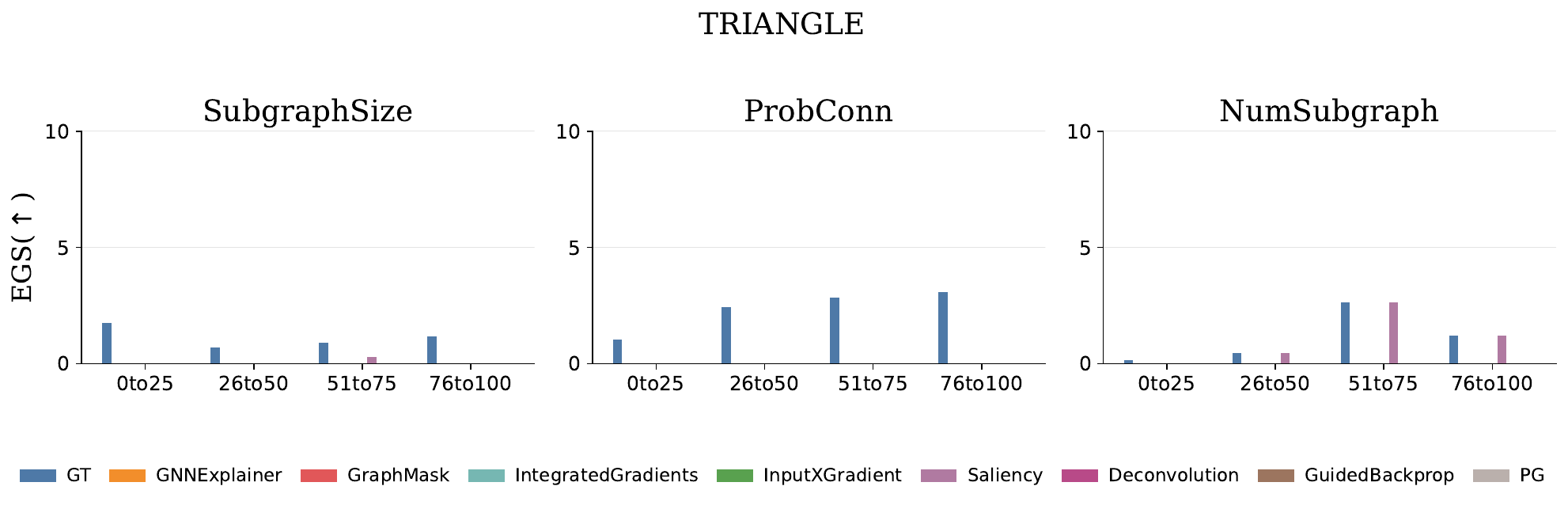}}

\caption{
\textbf{\egs comparison for ground truth and graph explainers using SAGE.} ~\looseness=-1 \egs results of explanation-guided training using ground-truth explanations and explanations generated from eight graph explainers on \tri over all graph properties. 
}
\end{center}
\vskip -0.2in
\end{figure*}

\begin{figure*}[h]
\vskip 0.2in
\begin{center}
\centerline{\includegraphics[width=\textwidth]{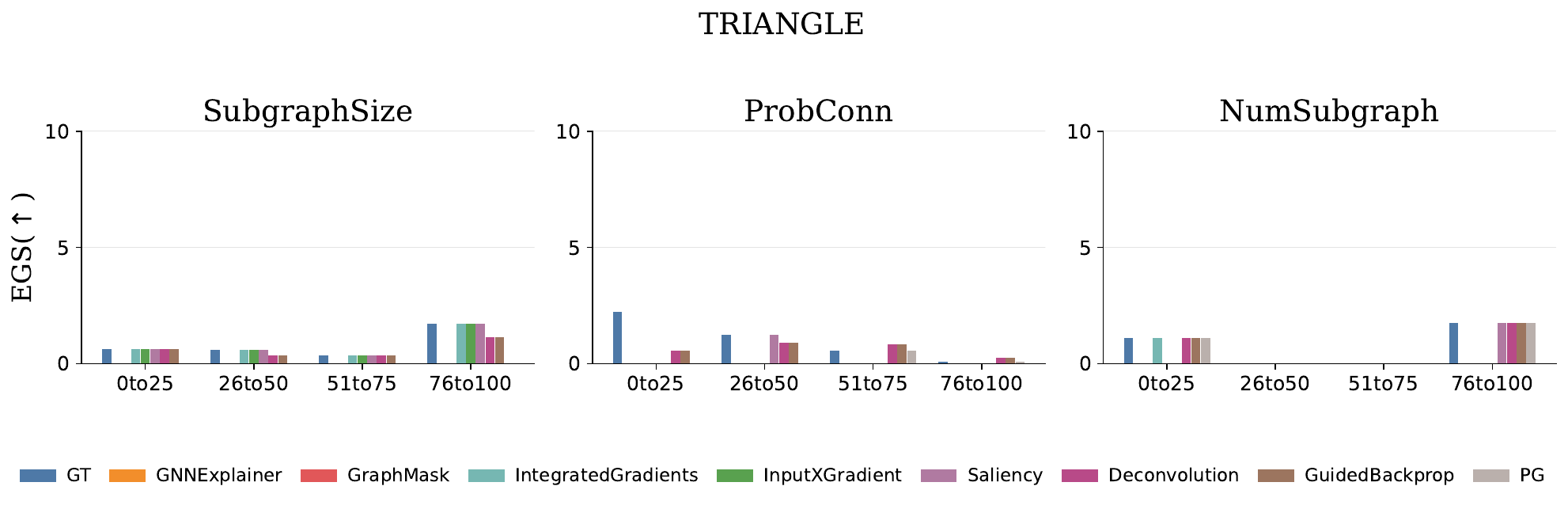}}

\caption{
\textbf{\egs comparison for ground truth and graph explainers using GIN.}  ~\looseness=-1 \egs results of explanation-guided training using ground-truth explanations and explanations generated from eight graph explainers on \tri over all graph properties. 
}
\end{center}
\vskip -0.2in
\end{figure*}

\begin{figure*}[h]
\vskip 0.2in
\begin{center}
\centerline{\includegraphics[width=\textwidth]{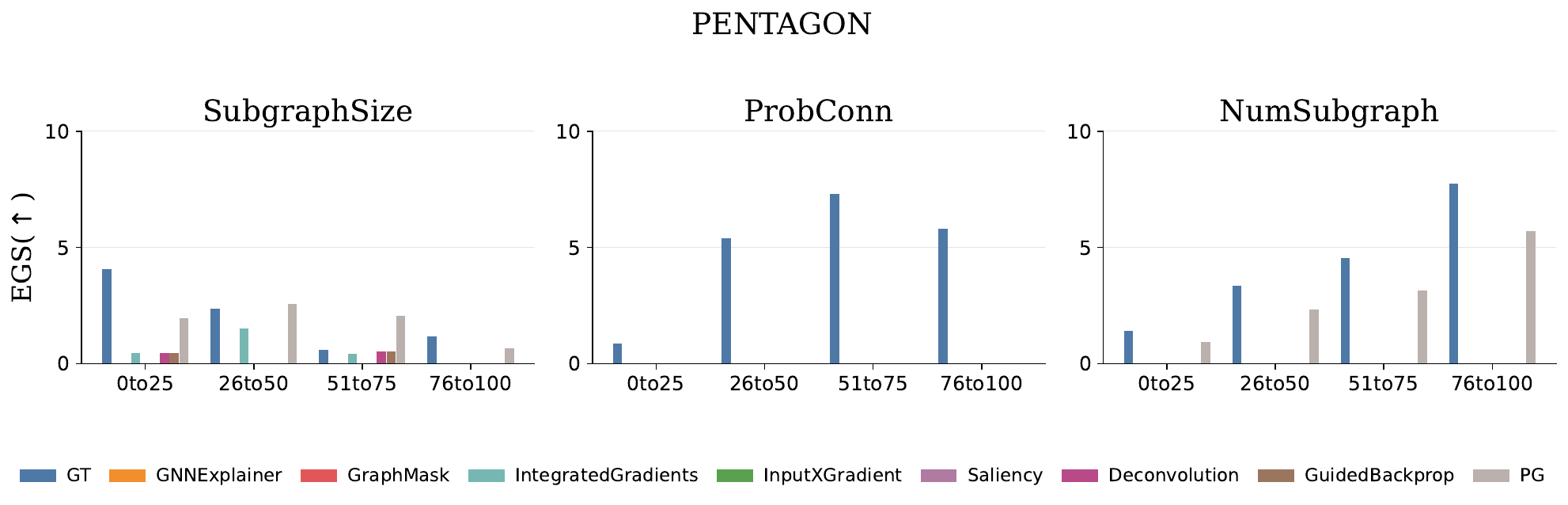}}

\caption{
\textbf{\egs comparison across ground truth and eight graph explainers using GAT.} ~\looseness=-1 \egs results of explanation-guided training using ground-truth explanations and explanations generated from eight graph explainers on \pen over all graph properties.  
}
\end{center}
\vskip -0.2in
\end{figure*}

\begin{figure*}[h]
\vskip 0.2in
\begin{center}
\centerline{\includegraphics[width=\textwidth]{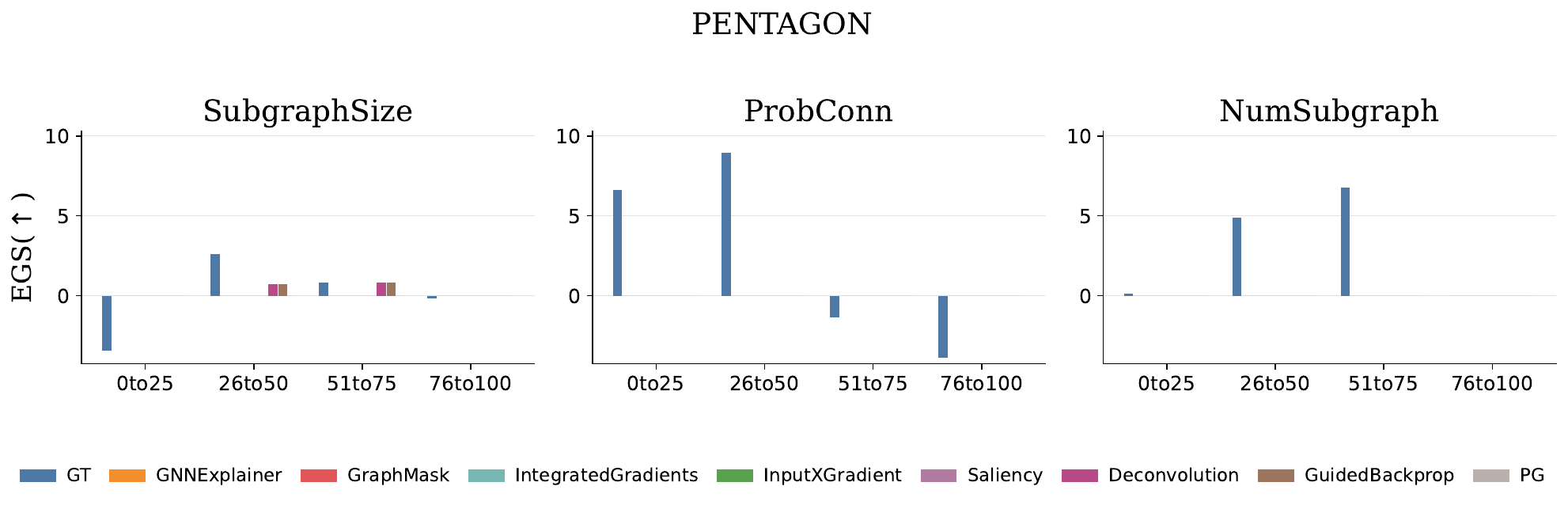}}

\caption{
\textbf{\egs comparison across ground truth and eight graph explainers using GCN.} ~\looseness=-1 \egs results of explanation-guided training using ground-truth explanations and explanations generated from eight graph explainers on \pen over all graph properties.  
}
\end{center}
\vskip -0.2in
\end{figure*} 

\begin{figure*}[h]
\vskip 0.2in
\begin{center}
\centerline{\includegraphics[width=\textwidth]{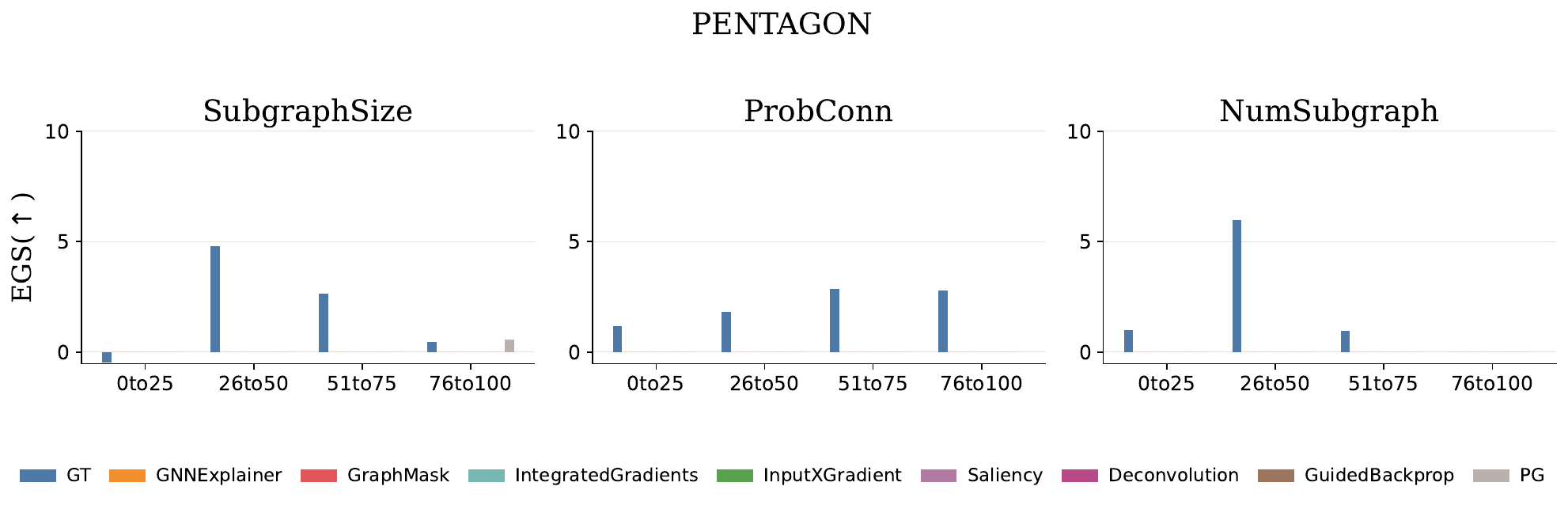}}

\caption{
\textbf{\egs comparison across ground truth and eight graph explainers using SAGE.} ~\looseness=-1 \egs results of explanation-guided training using ground-truth explanations and explanations generated from eight graph explainers on \pen over all graph properties.  
}
\end{center}
\vskip -0.2in
\end{figure*}

\begin{figure*}[h]
\vskip 0.2in
\begin{center}
\centerline{\includegraphics[width=\textwidth]{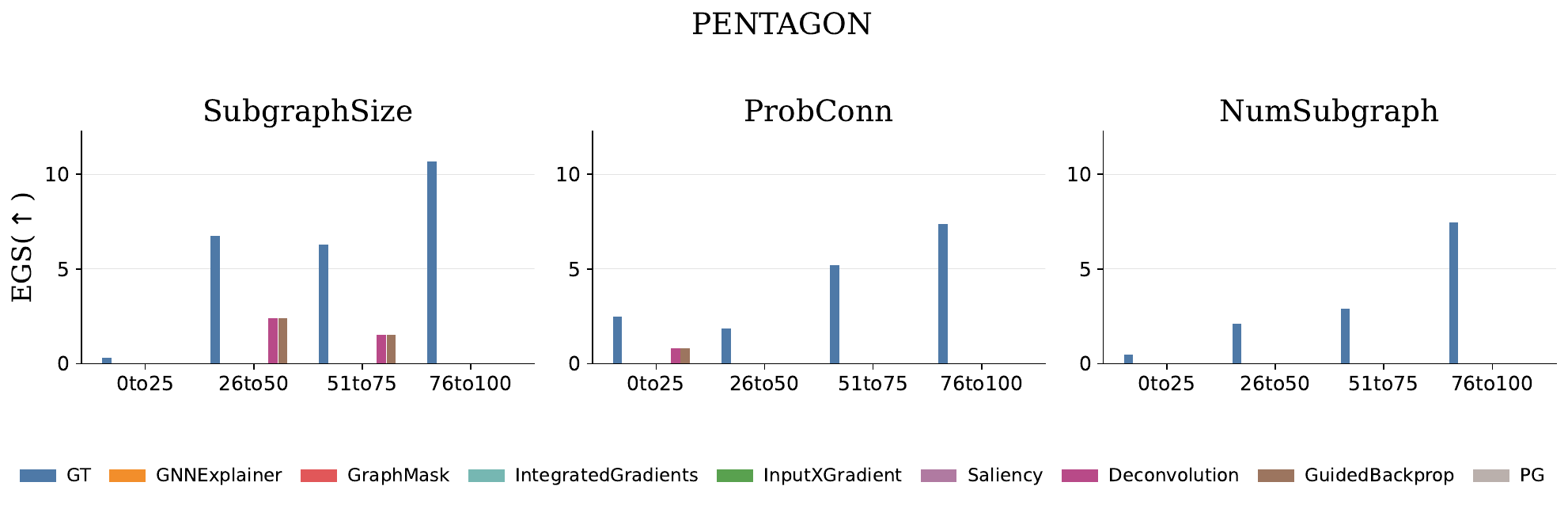}}

\caption{
\textbf{\egs comparison across ground truth and eight graph explainers using GIN.} ~\looseness=-1 \egs results of explanation-guided training using ground-truth explanations and explanations generated from eight graph explainers on \pen over all graph properties.  
}
\end{center}
\vskip -0.2in
\end{figure*}

\begin{figure*}[h]
\vskip 0.2in
\begin{center}
\centerline{\includegraphics[width=\textwidth]{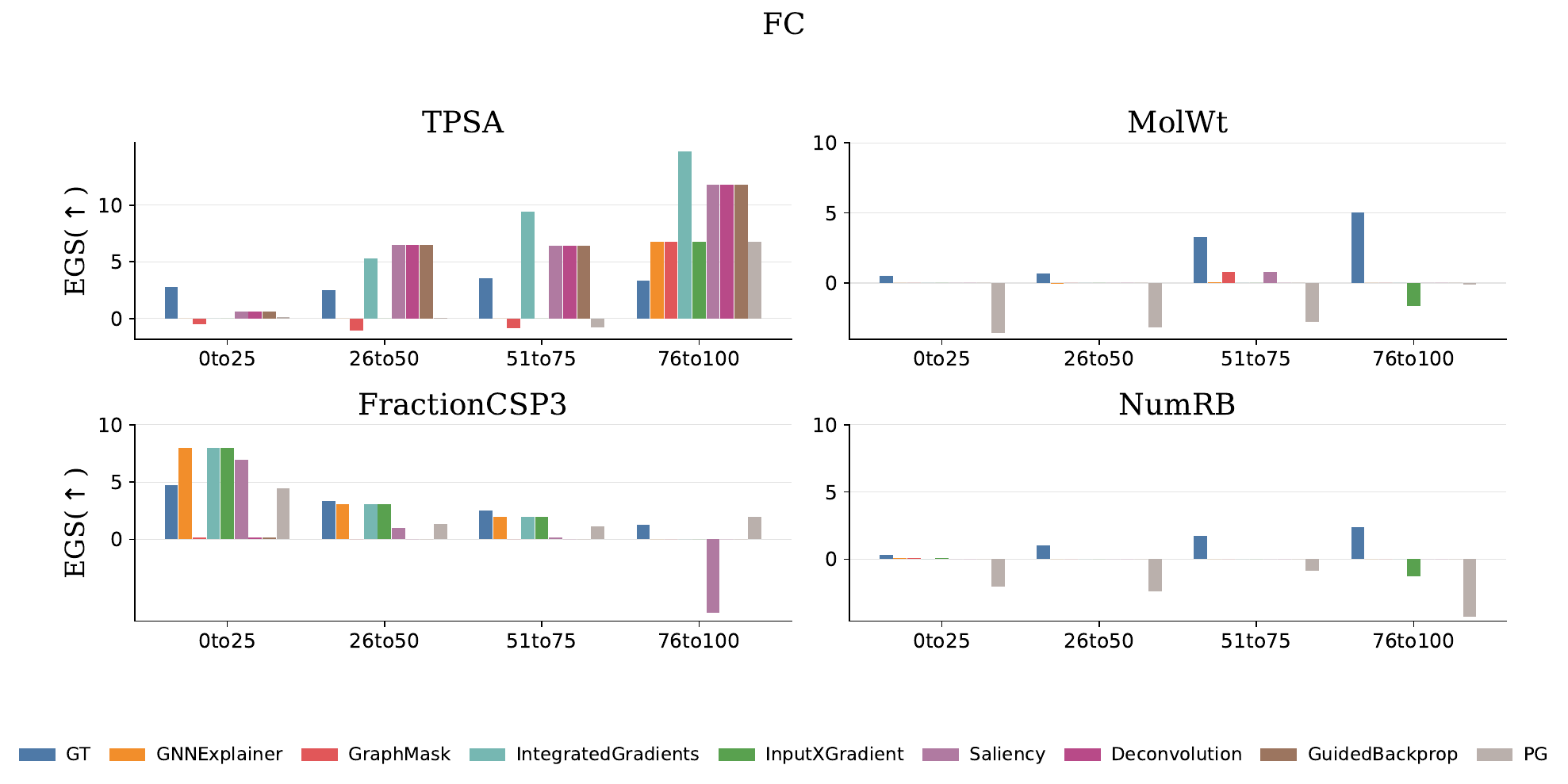}}

\caption{
\textbf{\egs comparison across ground truth and eight graph explainers using GAT.} ~\looseness=-1 \egs results of explanation-guided training using ground-truth explanations and explanations generated from eight graph explainers on \fc over all molecular properties.  
}
\end{center}
\vskip -0.2in
\end{figure*} 

\begin{figure*}[h]
\vskip 0.2in
\begin{center}
\centerline{\includegraphics[width=\textwidth]{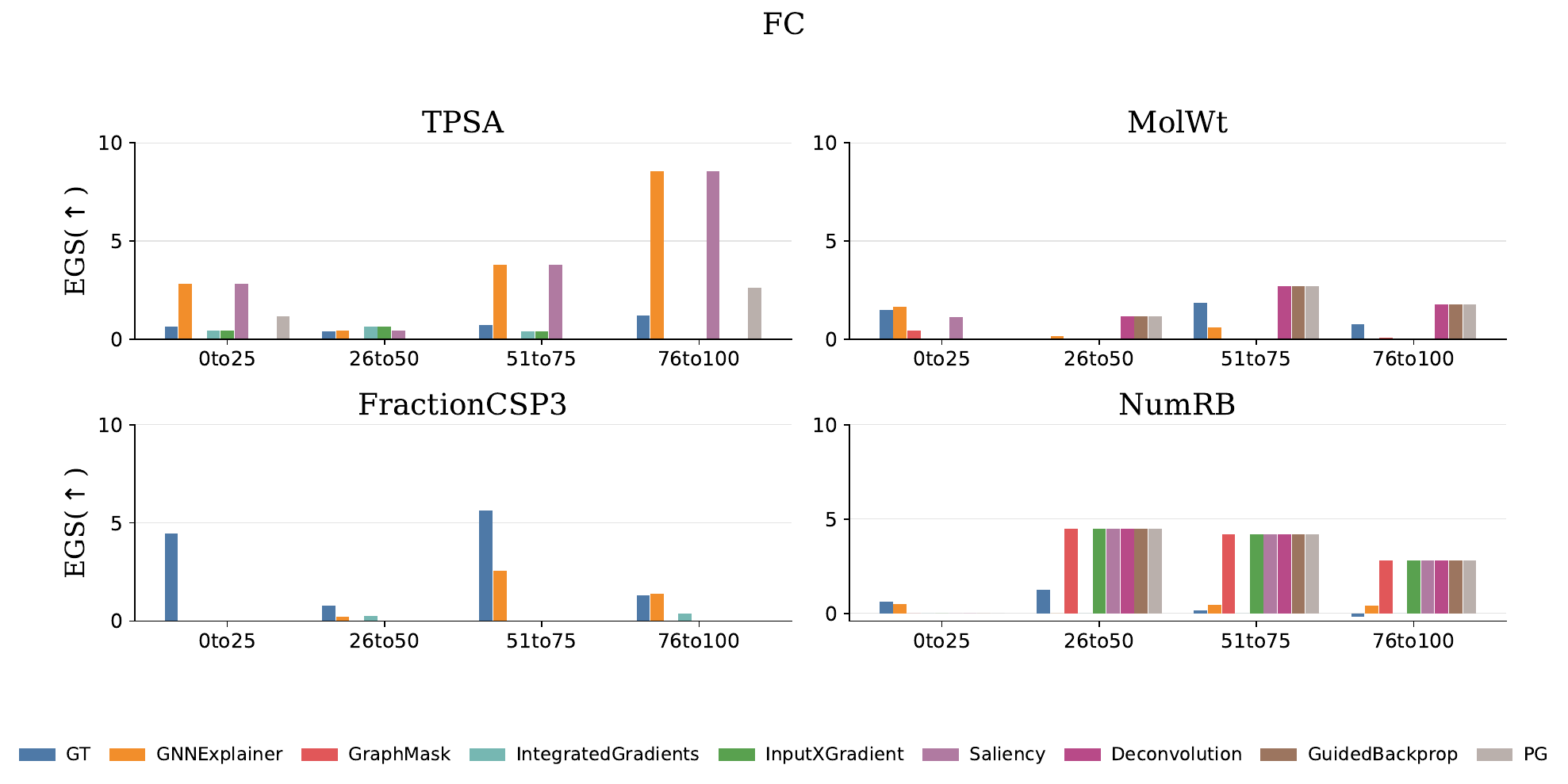}}

\caption{
\textbf{\egs comparison across ground truth and eight graph explainers using GCN.} ~\looseness=-1 \egs results of explanation-guided training using ground-truth explanations and explanations generated from eight graph explainers on \fc over all molecular properties.  
}
\end{center}
\vskip -0.2in
\end{figure*} 

\begin{figure*}[h]
\vskip 0.2in
\begin{center}
\centerline{\includegraphics[width=\textwidth]{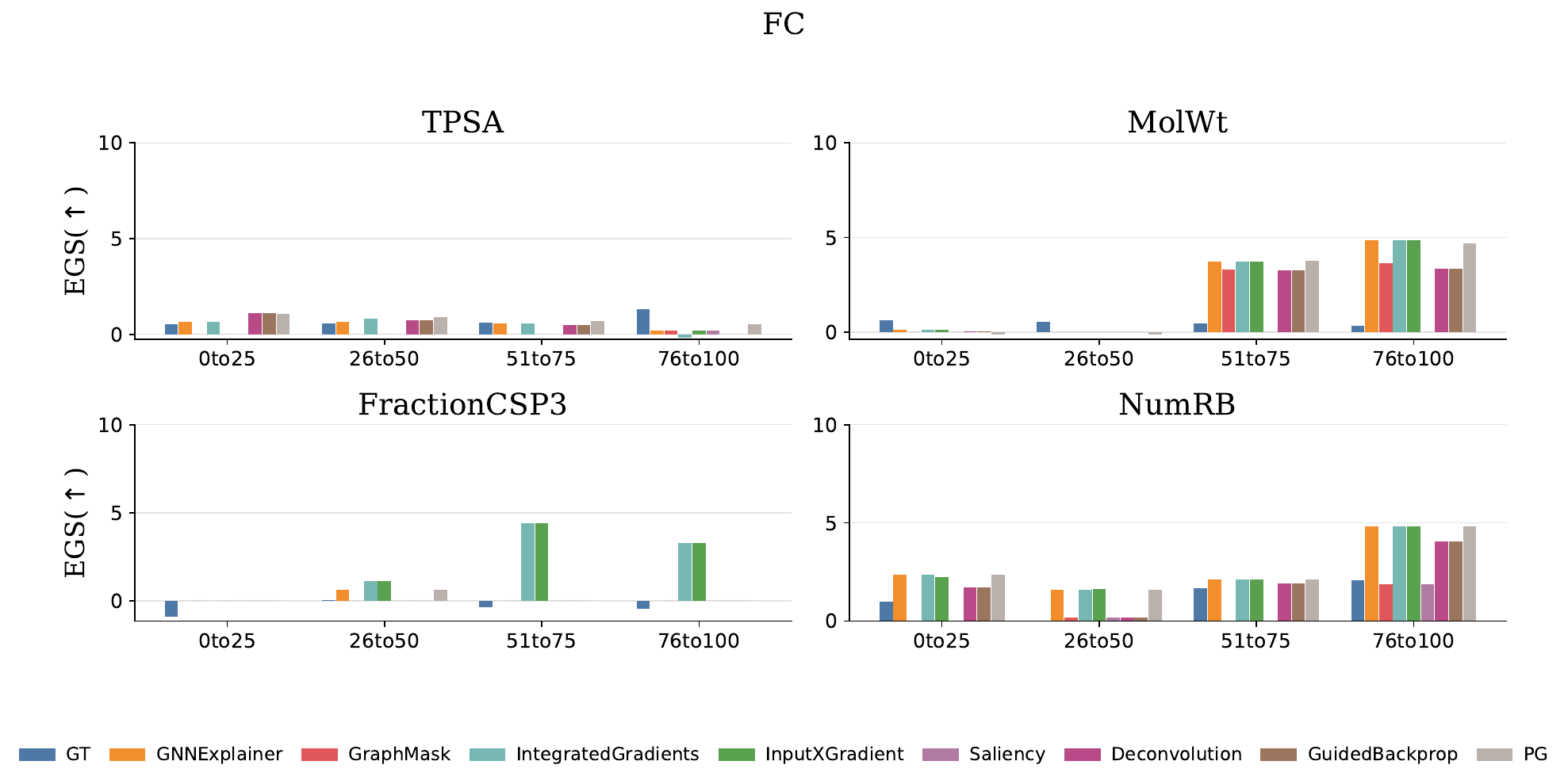}}

\caption{
\textbf{\egs comparison across ground truth and eight graph explainers using SAGE.} ~\looseness=-1 \egs results of explanation-guided training using ground-truth explanations and explanations generated from eight graph explainers on \fc over all molecular properties.  
}
\end{center}
\vskip -0.2in
\end{figure*} 

\begin{figure*}[h]
\vskip 0.2in
\begin{center}
\centerline{\includegraphics[width=\textwidth]{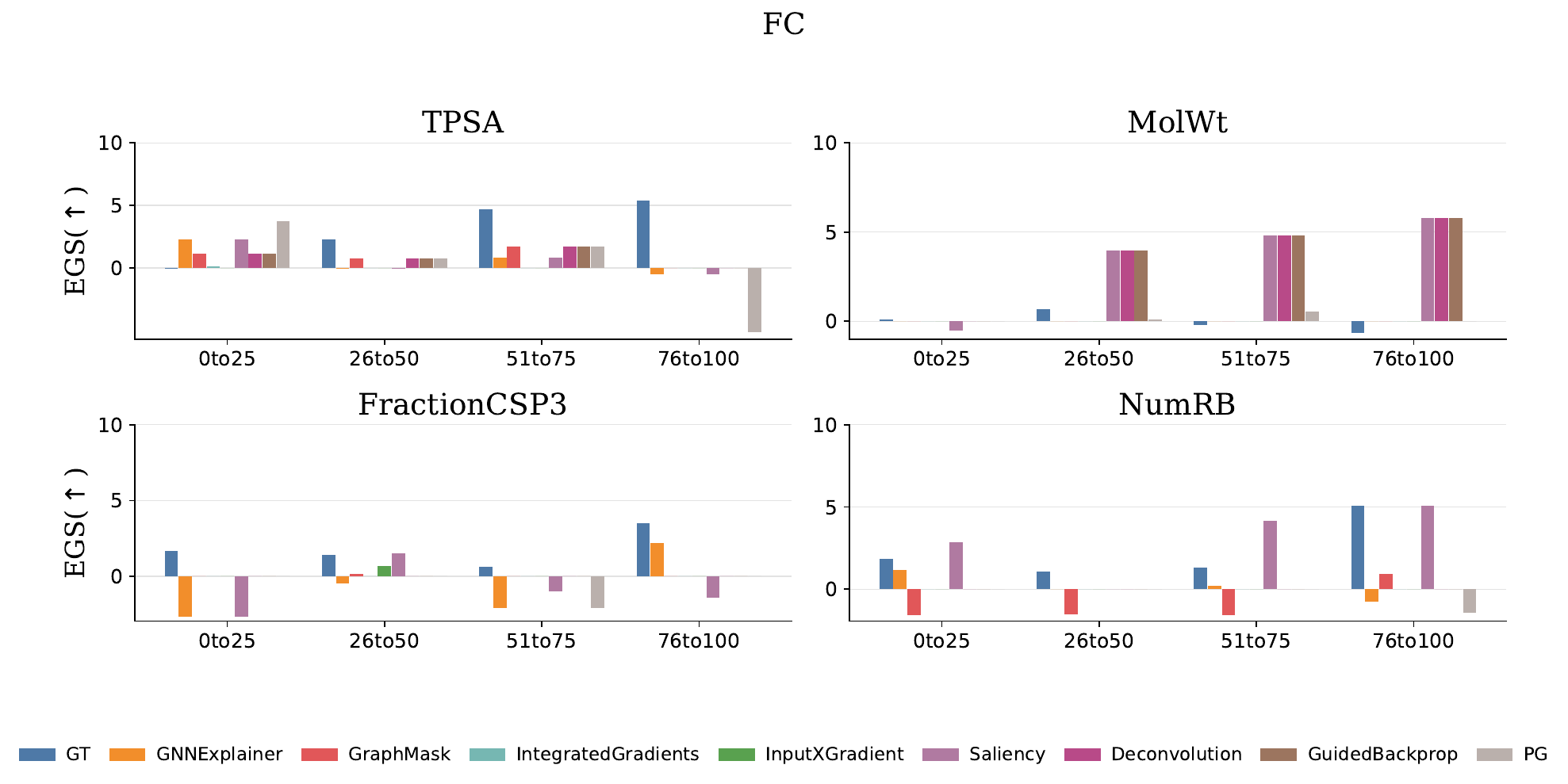}}

\caption{
\textbf{\egs comparison across ground truth and eight graph explainers using GIN.} ~\looseness=-1 \egs results of explanation-guided training using ground-truth explanations and explanations generated from eight graph explainers on \fc over all molecular properties.  
}
\end{center}
\vskip -0.2in
\end{figure*}

\begin{figure*}[h]
\vskip 0.2in
\begin{center}
\centerline{\includegraphics[width=\textwidth]{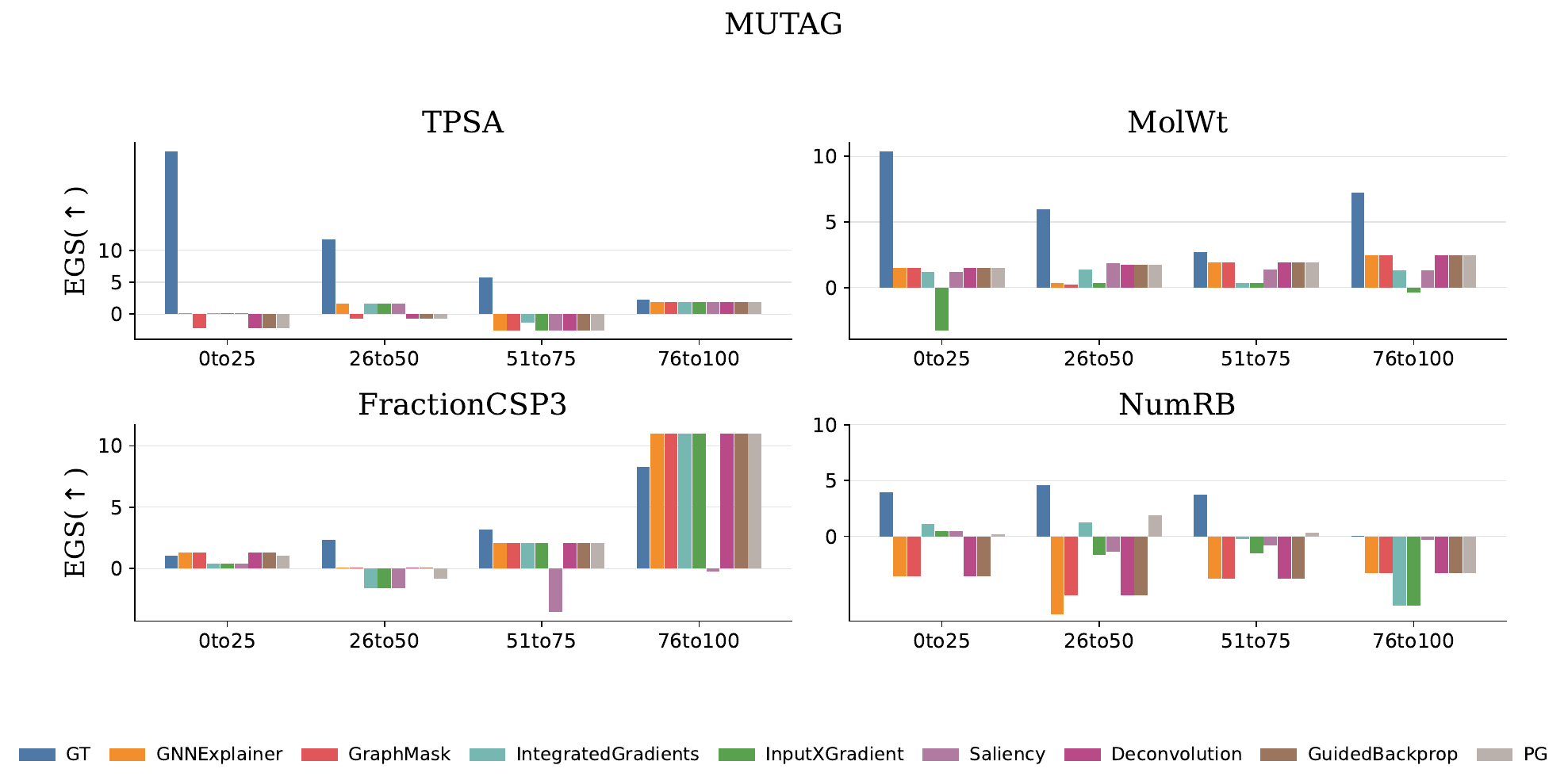}}

\caption{
\textbf{\egs comparison across ground truth and eight graph explainers using GAT.} ~\looseness=-1 \egs results of explanation-guided training using ground-truth explanations and explanations generated from eight graph explainers on \mutag over all molecular properties.  
}
\end{center}
\vskip -0.2in
\end{figure*}

\begin{figure*}[h]
\vskip 0.2in
\begin{center}
\centerline{\includegraphics[width=\textwidth]{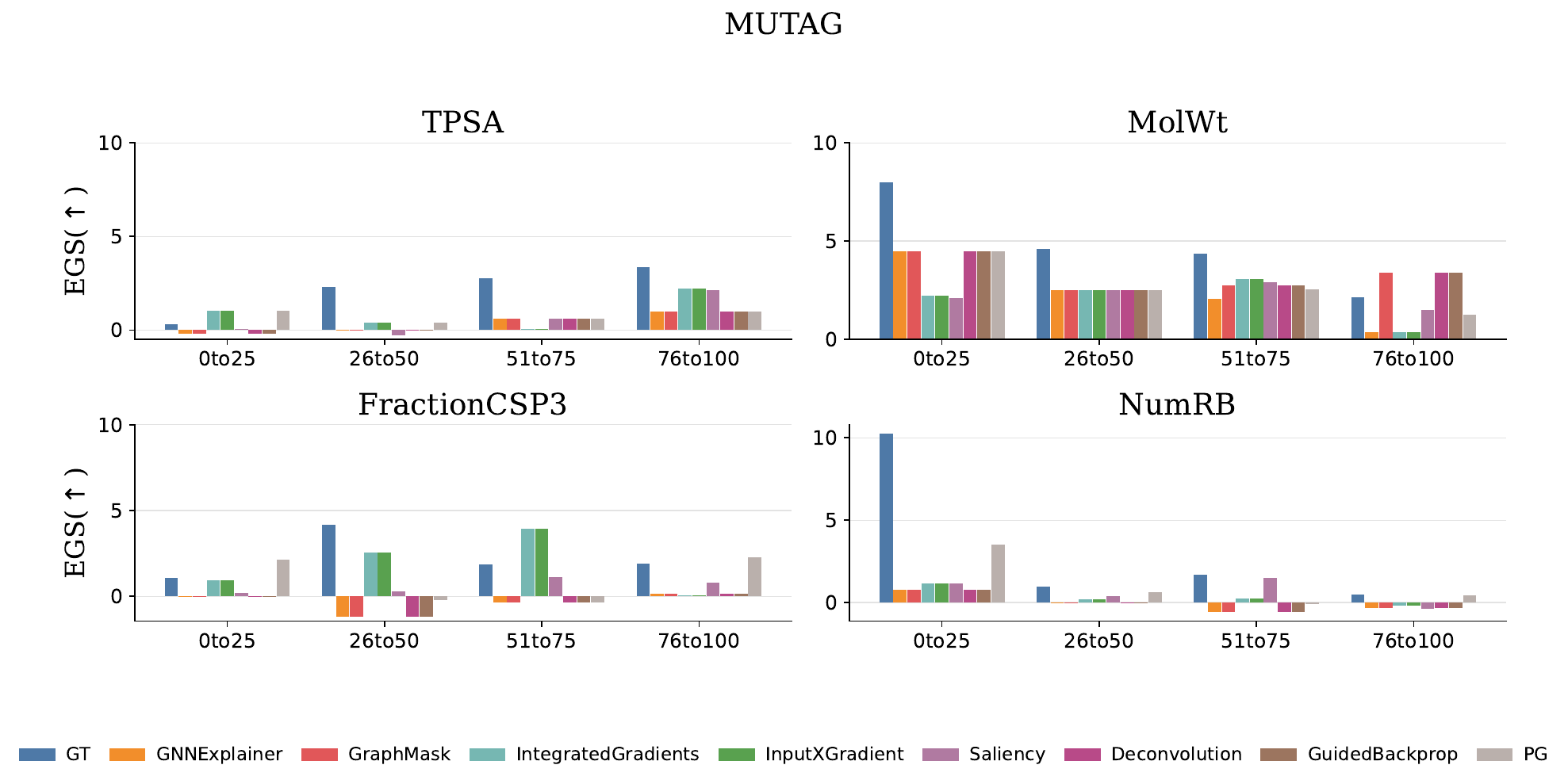}}

\caption{
\textbf{\egs comparison across ground truth and eight graph explainers using GCN.} ~\looseness=-1 \egs results of explanation-guided training using ground-truth explanations and explanations generated from eight graph explainers on \mutag over all molecular properties.  
}
\end{center}
\vskip -0.2in
\end{figure*}

\begin{figure*}[h]
\vskip 0.2in
\begin{center}
\centerline{\includegraphics[width=\textwidth]{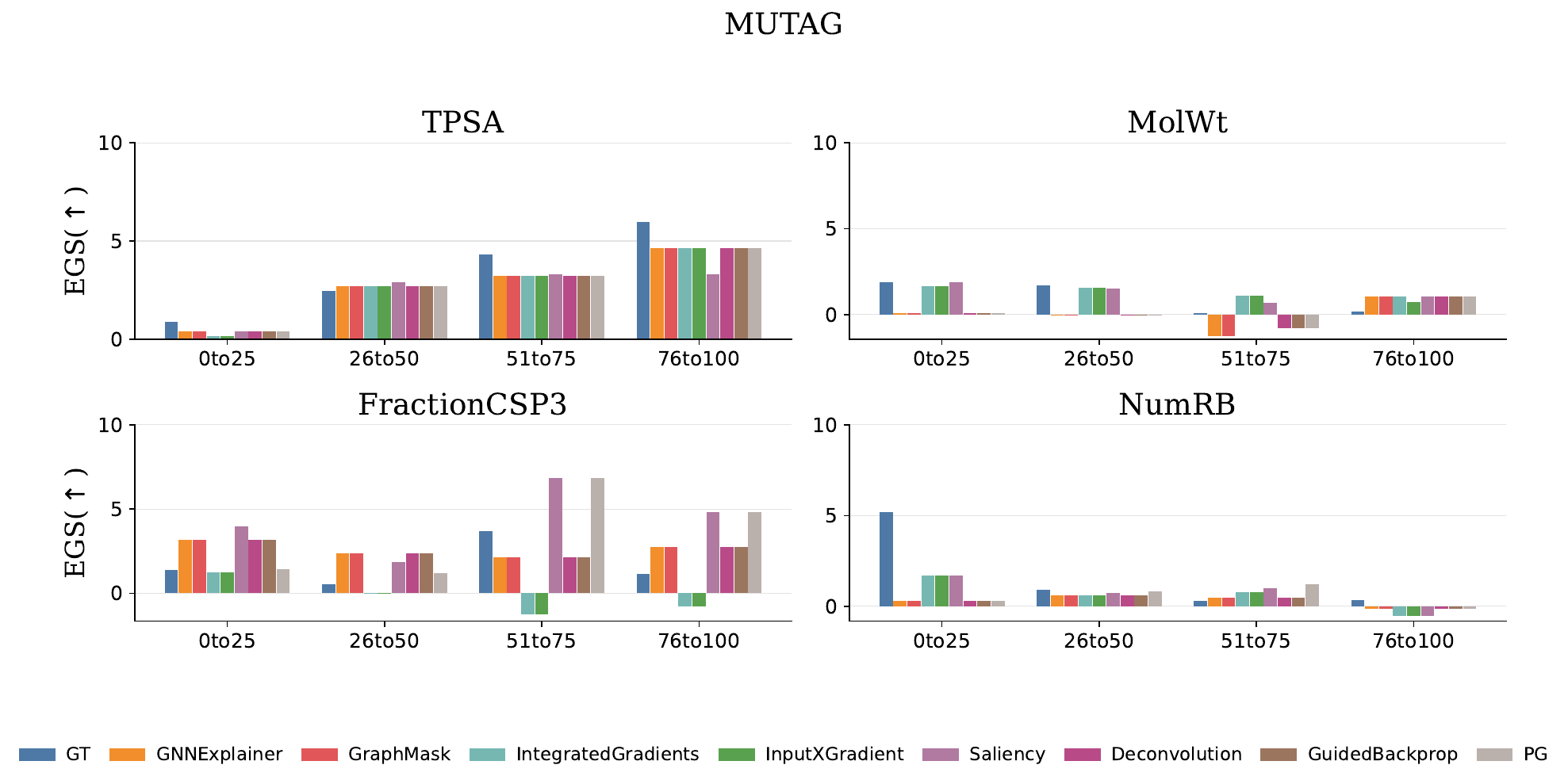}}

\caption{
\textbf{\egs comparison across ground truth and eight graph explainers using SAGE.} ~\looseness=-1 \egs results of explanation-guided training using ground-truth explanations and explanations generated from eight graph explainers on \mutag over all molecular properties.  
}
\end{center}
\vskip -0.2in
\end{figure*}

\begin{figure*}[h]
\vskip 0.2in
\begin{center}
\centerline{\includegraphics[width=\textwidth]{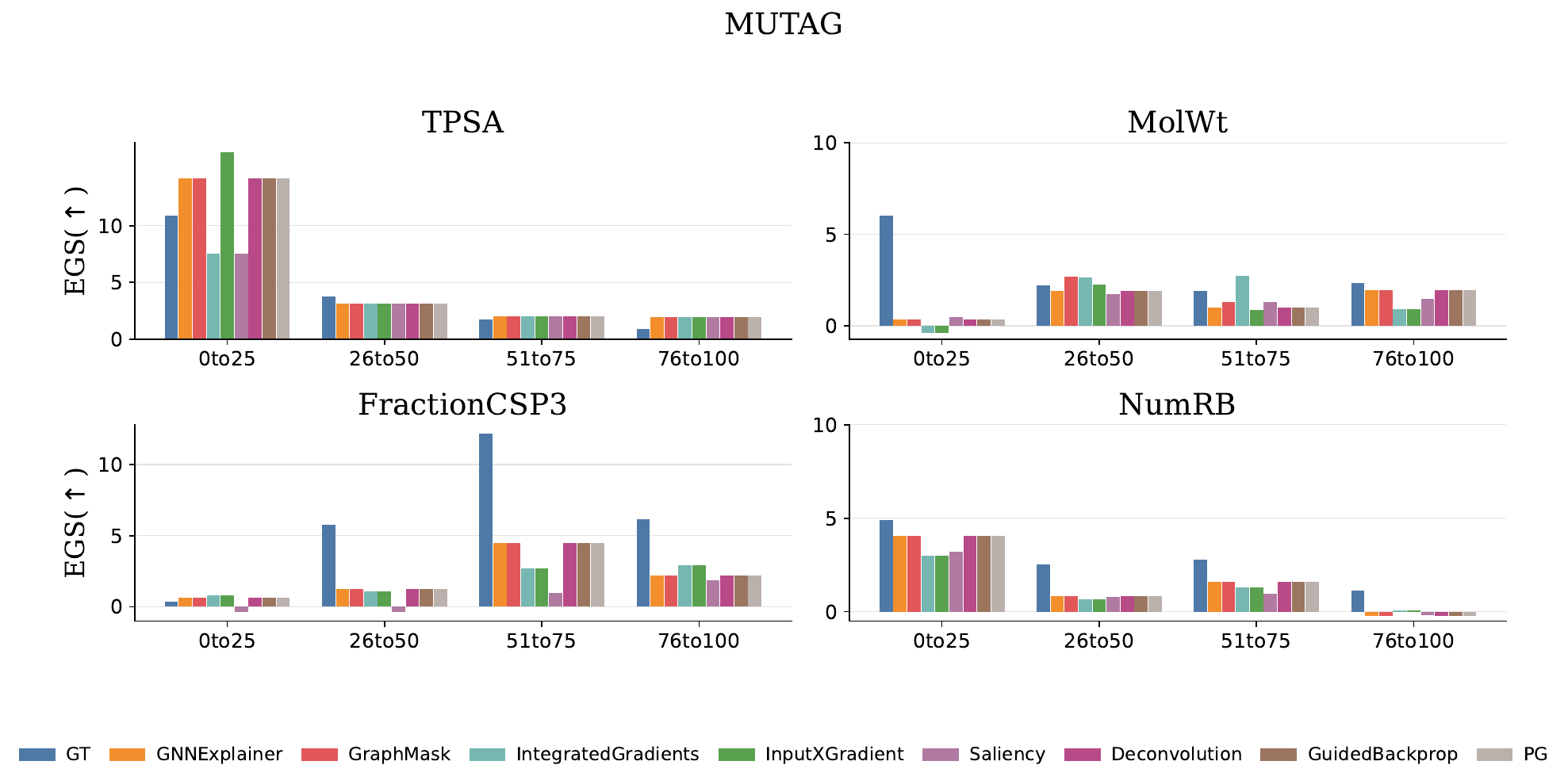}}

\caption{
\textbf{\egs comparison across ground truth and eight graph explainers using GIN.} ~\looseness=-1 \egs results of explanation-guided training using ground-truth explanations and explanations generated from eight graph explainers on \mutag over all molecular properties.  
}
\end{center}
\vskip -0.2in
\end{figure*}

\clearpage
\subsection{Tables}

\begin{table}[H]
    \centering
    \scriptsize
    \begin{tabular}{lccc}
    \toprule
        Property Type & OOD Testset & Model & IID BACC  \\ \midrule
        \multirow{8}{*}{MolWt} & \multirow{2}{*}{0to25} & GNN & 0.90 \std{0.05} \\ 
        ~ &  & EG-GNN & 0.89 \std{0.06} \\ \cline{2-4}
        ~ & \multirow{2}{*}{26to50} & GNN & 0.89 \std{0.07} \\ 
        ~ &  & EG-GNN & 0.93 \std{0.02} \\ \cline{2-4}
        ~ & \multirow{2}{*}{51to75} & GNN & 0.95 \std{0.02} \\ 
        ~ &  & EG-GNN & 0.94 \std{0.02} \\ \cline{2-4}
        ~ & \multirow{2}{*}{76to100} & GNN & 0.97 \std{0.03} \\ 
        ~ &  & EG-GNN & 0.96 \std{0.01} \\ \midrule
        \multirow{8}{*}{FractionCSP3} & \multirow{2}{*}{0to25} & GNN & 0.92 \std{0.04} \\ 
        &  & EG-GNN & 0.91 \std{0.03} \\ \cline{2-4}
         & \multirow{2}{*}{26to50} & GNN & 0.90 \std{0.09} \\ 
         &  & EG-GNN & 0.95 \std{0.02} \\ \cline{2-4}
         & \multirow{2}{*}{51to75} & GNN & 0.95 \std{0.02} \\ 
         &  & EG-GNN & 0.96 \std{0.01} \\ \cline{2-4}
         & \multirow{2}{*}{76to100} & GNN & 0.96 \std{0.02} \\ 
         &  & EG-GNN & 0.95 \std{0.01} \\ \midrule
        \multirow{8}{*}{NumRB} & \multirow{2}{*}{0to25} & GNN & 0.93 \std{0.02} \\ 
         &  & EG-GNN & 0.90 \std{0.05} \\ \cline{2-4}
         & \multirow{2}{*}{26to50} & GNN & 0.92 \std{0.05} \\ 
         &  & EG-GNN & 0.93 \std{0.05} \\ \cline{2-4}
         & \multirow{2}{*}{51to75} & GNN & 0.97 \std{0.01} \\ 
         &  & EG-GNN & 0.97 \std{0.02} \\ \cline{2-4}
        & \multirow{2}{*}{76to100} & GNN & 0.99 \std{0.01} \\ 
         &  & EG-GNN & 0.99 \std{0.01} \\ \midrule
        \multirow{8}{*}{TPSA} & \multirow{2}{*}{0to25} & GNN & 0.95 \std{0.05} \\ 
         &  & EG-GNN & 0.91 \std{0.06} \\ \cline{2-4}
         & \multirow{2}{*}{26to50} & GNN & 0.99 \std{0.01} \\ 
         &  & EG-GNN & 0.99 \std{0.01} \\ \cline{2-4}
         & \multirow{2}{*}{51to75} & GNN & 0.95 \std{0.02} \\ 
         &  & EG-GNN & 0.95 \std{0.04} \\ \cline{2-4}
         & \multirow{2}{*}{76to100} & GNN & 0.94 \std{0.03} \\ 
         &  & EG-GNN & 0.95 \std{0.01} \\
         \bottomrule
    \end{tabular}
    
    \caption{\textbf{Performance results for \fc using ground-truth explanations in EG-GNNs using GCN.} ~\looseness=-1 Comparison of baseline GNNs and EG-GNNs across OOD test splits and molecular properties. The table reports IID balanced accuracy (IID BACC). Results are presented as mean $\pm$ standard deviation across cross-validation folds for the IID setting.}
    \label{tab:egs_gte_fc}
\end{table}
\clearpage

\begin{table}[H]
    \centering
    \scriptsize
    \begin{tabular}{lccc}
    \toprule
        Property Type & OOD Testset & Model & IID BACC \\ \midrule
        \multirow{8}{*}{MolWt} & \multirow{2}{*}{0to25} & GNN & 0.92 \std{0.08} \\ 
        ~ &  & EG-GNN & 0.93 \std{0.06} \\ \cline{2-4}
        ~ & \multirow{2}{*}{26to50} & GNN & 0.94 \std{0.04} \\ 
        ~ &  & EG-GNN & 0.97 \std{0.02} \\ \cline{2-4}
        ~ & \multirow{2}{*}{51to75} & GNN & 0.96 \std{0.03} \\ 
        ~ &  & EG-GNN & 0.96 \std{0.03} \\ \cline{2-4}
        ~ & \multirow{2}{*}{76to100} & GNN & 0.94 \std{0.03} \\ 
        ~ &  & EG-GNN & 0.96 \std{0.02} \\ \midrule
        \multirow{8}{*}{FractionCSP3} & \multirow{2}{*}{0to25} & GNN & 0.93 \std{0.04} \\ 
        &  & EG-GNN & 0.95 \std{0.02} \\ \cline{2-4}
         & \multirow{2}{*}{26to50} & GNN & 0.93 \std{0.02} \\ 
         &  & EG-GNN & 0.92 \std{0.05} \\ \cline{2-4}
         & \multirow{2}{*}{51to75} & GNN & 0.95 \std{0.02} \\ 
         &  & EG-GNN & 0.95 \std{0.03} \\ \cline{2-4}
         & \multirow{2}{*}{76to100} & GNN & 0.92 \std{0.05} \\ 
         &  & EG-GNN & 0.93 \std{0.05} \\ \midrule
        \multirow{8}{*}{NumRB} & \multirow{2}{*}{0to25} & GNN & 0.92 \std{0.05} \\ 
         &  & EG-GNN & 0.92 \std{0.02} \\ \cline{2-4}
         & \multirow{2}{*}{26to50} & GNN & 0.93 \std{0.04} \\ 
         &  & EG-GNN & 0.93 \std{0.03} \\ \cline{2-4}
         & \multirow{2}{*}{51to75} & GNN & 0.96 \std{0.02} \\ 
         &  & EG-GNN & 0.97 \std{0.01} \\ \cline{2-4}
        & \multirow{2}{*}{76to100} & GNN & 0.97 \std{0.03} \\ 
         &  & EG-GNN & 0.98 \std{0.02} \\ \midrule
        \multirow{8}{*}{TPSA} & \multirow{2}{*}{0to25} & GNN & 0.90 \std{0.08} \\ 
         &  & EG-GNN & 0.95 \std{0.03} \\ \cline{2-4}
         & \multirow{2}{*}{26to50} & GNN & 0.98 \std{0.01} \\ 
         &  & EG-GNN & 0.98 \std{0.02} \\ \cline{2-4}
         & \multirow{2}{*}{51to75} & GNN & 0.95 \std{0.02} \\ 
         &  & EG-GNN & 0.95 \std{0.02} \\ \cline{2-4}
         & \multirow{2}{*}{76to100} & GNN & 0.97 \std{0.01} \\ 
         &  & EG-GNN & 0.96 \std{0.01} \\
         \bottomrule
    \end{tabular}
    
    \caption{\textbf{Performance results for \fc using ground-truth explanations in EG-GNNs using GAT.} ~\looseness=-1 Comparison of baseline GNNs and EG-GNNs across OOD test splits and molecular properties. The table reports IID balanced accuracy (IID BACC). Results are presented as mean $\pm$ standard deviation across cross-validation folds for the IID setting.}
    \label{tab:egs_gte_fc}
\end{table}

\begin{table}[H]
    \centering
    \scriptsize
    \begin{tabular}{lccc}
    \toprule
        Property Type & OOD Testset & Model & IID BACC \\ \midrule
        \multirow{8}{*}{MolWt} & \multirow{2}{*}{0to25} & GNN & 0.86 \std{0.07} \\ 
        ~ &  & EG-GNN & 0.85 \std{0.10} \\ \cline{2-4}
        ~ & \multirow{2}{*}{26to50} & GNN & 0.92 \std{0.02} \\ 
        ~ &  & EG-GNN & 0.92 \std{0.02} \\ \cline{2-4}
        ~ & \multirow{2}{*}{51to75} & GNN & 0.94 \std{0.02} \\ 
        ~ &  & EG-GNN & 0.95 \std{0.02} \\ \cline{2-4}
        ~ & \multirow{2}{*}{76to100} & GNN & 0.95 \std{0.02} \\ 
        ~ &  & EG-GNN & 0.95 \std{0.02} \\ \midrule
        \multirow{8}{*}{FractionCSP3} & \multirow{2}{*}{0to25} & GNN & 0.91 \std{0.05} \\ 
        &  & EG-GNN & 0.92 \std{0.04} \\ \cline{2-4}
         & \multirow{2}{*}{26to50} & GNN & 0.93 \std{0.03} \\ 
         &  & EG-GNN & 0.91 \std{0.06} \\ \cline{2-4}
         & \multirow{2}{*}{51to75} & GNN & 0.92 \std{0.03} \\ 
         &  & EG-GNN & 0.91 \std{0.05} \\ \cline{2-4}
         & \multirow{2}{*}{76to100} & GNN & 0.91 \std{0.02} \\ 
         &  & EG-GNN & 0.92 \std{0.02} \\ \midrule
        \multirow{8}{*}{NumRB} & \multirow{2}{*}{0to25} & GNN & 0.92 \std{0.02} \\ 
         &  & EG-GNN & 0.91 \std{0.04} \\ \cline{2-4}
         & \multirow{2}{*}{26to50} & GNN & 0.93 \std{0.04} \\ 
         &  & EG-GNN & 0.94 \std{0.04} \\ \cline{2-4}
         & \multirow{2}{*}{51to75} & GNN & 0.95 \std{0.02} \\ 
         &  & EG-GNN & 0.94 \std{0.02} \\ \cline{2-4}
        & \multirow{2}{*}{76to100} & GNN & 0.95 \std{0.04} \\ 
         &  & EG-GNN & 0.94 \std{0.02} \\ \midrule
        \multirow{8}{*}{TPSA} & \multirow{2}{*}{0to25} & GNN & 0.89 \std{0.07} \\ 
         &  & EG-GNN & 0.86 \std{0.07} \\ \cline{2-4}
         & \multirow{2}{*}{26to50} & GNN & 0.97 \std{0.02} \\ 
         &  & EG-GNN & 0.98 \std{0.02} \\ \cline{2-4}
         & \multirow{2}{*}{51to75} & GNN & 0.94 \std{0.03} \\ 
         &  & EG-GNN & 0.92 \std{0.03} \\ \cline{2-4}
         & \multirow{2}{*}{76to100} & GNN & 0.94 \std{0.04} \\ 
         &  & EG-GNN & 0.95 \std{0.02} \\
         \bottomrule
    \end{tabular}
    
    \caption{\textbf{Performance results for \fc using ground-truth explanations in EG-GNNs using SAGE.} ~\looseness=-1 Comparison of baseline GNNs and EG-GNNs across OOD test splits and molecular properties. The table reports IID balanced accuracy (IID BACC). Results are presented as mean $\pm$ standard deviation across cross-validation folds for the IID setting.}
    \label{tab:egs_gte_fc}
\end{table}

\begin{table}[H]
    \centering
    \scriptsize
    \begin{tabular}{lccc}
    \toprule
        Property Type & OOD Testset & Model & IID BACC \\ \midrule
        \multirow{8}{*}{MolWt} & \multirow{2}{*}{0to25} & GNN & 0.89 \std{0.06} \\ 
        ~ &  & EG-GNN & 0.89 \std{0.06} \\ \cline{2-4}
        ~ & \multirow{2}{*}{26to50} & GNN & 0.95 \std{0.01} \\ 
        ~ &  & EG-GNN & 0.93 \std{0.05} \\ \cline{2-4}
        ~ & \multirow{2}{*}{51to75} & GNN & 0.94 \std{0.02} \\ 
        ~ &  & EG-GNN & 0.93 \std{0.02} \\ \cline{2-4}
        ~ & \multirow{2}{*}{76to100} & GNN & 0.95 \std{0.02} \\ 
        ~ &  & EG-GNN & 0.95 \std{0.02} \\ \midrule
        \multirow{8}{*}{FractionCSP3} & \multirow{2}{*}{0to25} & GNN & 0.91 \std{0.06} \\ 
        &  & EG-GNN & 0.92 \std{0.04} \\ \cline{2-4}
         & \multirow{2}{*}{26to50} & GNN & 0.94 \std{0.02} \\ 
         &  & EG-GNN & 0.96 \std{0.02} \\ \cline{2-4}
         & \multirow{2}{*}{51to75} & GNN & 0.94 \std{0.03} \\ 
         &  & EG-GNN & 0.94 \std{0.02} \\ \cline{2-4}
         & \multirow{2}{*}{76to100} & GNN & 0.95 \std{0.03} \\ 
         &  & EG-GNN & 0.96 \std{0.01} \\ \midrule
        \multirow{8}{*}{NumRB} & \multirow{2}{*}{0to25} & GNN & 0.92 \std{0.03} \\ 
         &  & EG-GNN & 0.90 \std{0.03} \\ \cline{2-4}
         & \multirow{2}{*}{26to50} & GNN & 0.93 \std{0.04} \\ 
         &  & EG-GNN & 0.95 \std{0.03} \\ \cline{2-4}
         & \multirow{2}{*}{51to75} & GNN & 0.94 \std{0.03} \\ 
         &  & EG-GNN & 0.96 \std{0.01} \\ \cline{2-4}
        & \multirow{2}{*}{76to100} & GNN & 0.96 \std{0.02} \\ 
         &  & EG-GNN & 0.98 \std{0.02} \\ \midrule
        \multirow{8}{*}{TPSA} & \multirow{2}{*}{0to25} & GNN & 0.93 \std{0.04} \\ 
         &  & EG-GNN & 0.94 \std{0.04} \\ \cline{2-4}
         & \multirow{2}{*}{26to50} & GNN & 0.99 \std{0.01} \\ 
         &  & EG-GNN & 0.99 \std{0.01} \\ \cline{2-4}
         & \multirow{2}{*}{51to75} & GNN & 0.96 \std{0.02} \\ 
         &  & EG-GNN & 0.94 \std{0.04} \\ \cline{2-4}
         & \multirow{2}{*}{76to100} & GNN & 0.96 \std{0.01} \\ 
         &  & EG-GNN & 0.98 \std{0.01} \\
         \bottomrule
    \end{tabular}
    
    \caption{\textbf{Performance results for \fc using ground-truth explanations in EG-GNNs using GIN.} ~\looseness=-1 Comparison of baseline GNNs and EG-GNNs across OOD test splits and molecular properties. The table reports IID balanced accuracy (IID BACC). Results are presented as mean $\pm$ standard deviation across cross-validation folds for the IID setting.}
    \label{tab:egs_gte_fc}
\end{table}

\begin{table}[H]
    \centering
    \scriptsize
    \begin{tabular}{lccc}
    \toprule
        Property Type & OOD Testset & Model & IID BACC \\ \midrule
        \multirow{8}{*}{MolWt} & \multirow{2}{*}{0to25} & GNN & 0.98 \std{0.01} \\ 
        ~ &  & EG-GNN & 0.98 \std{0.01} \\ \cline{2-4}
        ~ & \multirow{2}{*}{26to50} & GNN & 0.96 \std{0.03} \\ 
        ~ &  & EG-GNN & 0.97 \std{0.01} \\ \cline{2-4}
        ~ & \multirow{2}{*}{51to75} & GNN & 0.97 \std{0.02} \\ 
        ~ &  & EG-GNN & 0.96 \std{0.01} \\ \cline{2-4}
        ~ & \multirow{2}{*}{76to100} & GNN & 0.93 \std{0.01} \\ 
        ~ &  & EG-GNN & 0.95 \std{0.01} \\ \midrule
        \multirow{8}{*}{FractionCSP3} & \multirow{2}{*}{0to25} & GNN & 0.97 \std{0.01} \\ 
        &  & EG-GNN & 0.98 \std{0.01} \\ \cline{2-4}
         & \multirow{2}{*}{26to50} & GNN & 0.96 \std{0.02} \\ 
         &  & EG-GNN & 0.97 \std{0.01} \\ \cline{2-4}
         & \multirow{2}{*}{51to75} & GNN & 0.98 \std{0.01} \\ 
         &  & EG-GNN & 0.97 \std{0.03} \\ \cline{2-4}
         & \multirow{2}{*}{76to100} & GNN & 0.97 \std{0.01} \\ 
         &  & EG-GNN & 0.97 \std{0.01} \\ \midrule
        \multirow{8}{*}{NumRB} & \multirow{2}{*}{0to25} & GNN & 0.97 \std{0.01} \\ 
         &  & EG-GNN & 0.97 \std{0.03} \\ \cline{2-4}
         & \multirow{2}{*}{26to50} & GNN & 0.96 \std{0.02} \\ 
         &  & EG-GNN & 0.97 \std{0.01} \\ \cline{2-4}
         & \multirow{2}{*}{51to75} & GNN & 0.98 \std{0.01} \\ 
         &  & EG-GNN & 0.98 \std{0.01} \\ \cline{2-4}
        & \multirow{2}{*}{76to100} & GNN & 0.98 \std{0.01} \\ 
         &  & EG-GNN & 0.98 \std{0.02} \\ \midrule
        \multirow{8}{*}{TPSA} & \multirow{2}{*}{0to25} & GNN & 0.98 \std{0.01} \\ 
         &  & EG-GNN & 0.98 \std{0.01} \\ \cline{2-4}
         & \multirow{2}{*}{26to50} & GNN & 0.97 \std{0.01} \\ 
         &  & EG-GNN & 0.97 \std{0.01} \\ \cline{2-4}
         & \multirow{2}{*}{51to75} & GNN & 0.97 \std{0.01} \\ 
         &  & EG-GNN & 0.97 \std{0.02} \\ \cline{2-4}
         & \multirow{2}{*}{76to100} & GNN & 0.97 \std{0.01} \\ 
         &  & EG-GNN & 0.97 \std{0.01} \\
         \bottomrule
    \end{tabular}
    
    \caption{\textbf{Performance results for \mutag using ground-truth explanations in EG-GNNs using GCN.} ~\looseness=-1 Comparison of baseline GNNs and EG-GNNs across OOD test splits and molecular properties. The table reports IID balanced accuracy (IID BACC). Results are presented as mean $\pm$ standard deviation across cross-validation folds for the IID setting.}
    \label{tab:egs_gte_fc}
\end{table}

\begin{table}[H]
    \centering
    \scriptsize
    \begin{tabular}{lccc}
    \toprule
        Property Type & OOD Testset & Model & IID BACC \\ \midrule
        \multirow{8}{*}{MolWt} & \multirow{2}{*}{0to25} & GNN & 0.97 \std{0.01} \\ 
        ~ &  & EG-GNN & 0.97 \std{0.01} \\ \cline{2-4}
        ~ & \multirow{2}{*}{26to50} & GNN & 0.95 \std{0.02} \\ 
        ~ &  & EG-GNN & 0.95 \std{0.02} \\ \cline{2-4}
        ~ & \multirow{2}{*}{51to75} & GNN & 0.92 \std{0.02} \\ 
        ~ &  & EG-GNN & 0.95 \std{0.01} \\ \cline{2-4}
        ~ & \multirow{2}{*}{76to100} & GNN & 0.93 \std{0.02} \\ 
        ~ &  & EG-GNN & 0.91 \std{0.05} \\ \midrule
        \multirow{8}{*}{FractionCSP3} & \multirow{2}{*}{0to25} & GNN & 0.94 \std{0.02} \\ 
        &  & EG-GNN & 0.96 \std{0.02} \\ \cline{2-4}
         & \multirow{2}{*}{26to50} & GNN & 0.94 \std{0.02} \\ 
         &  & EG-GNN & 0.93 \std{0.03} \\ \cline{2-4}
         & \multirow{2}{*}{51to75} & GNN & 0.95 \std{0.03} \\ 
         &  & EG-GNN & 0.96 \std{0.02} \\ \cline{2-4}
         & \multirow{2}{*}{76to100} & GNN & 0.96 \std{0.01} \\ 
         &  & EG-GNN & 0.96 \std{0.02} \\ \midrule
        \multirow{8}{*}{NumRB} & \multirow{2}{*}{0to25} & GNN & 0.96 \std{0.01} \\ 
         &  & EG-GNN & 0.96 \std{0.01} \\ \cline{2-4}
         & \multirow{2}{*}{26to50} & GNN & 0.94 \std{0.01} \\ 
         &  & EG-GNN & 0.95 \std{0.01} \\ \cline{2-4}
         & \multirow{2}{*}{51to75} & GNN & 0.98 \std{0.01} \\ 
         &  & EG-GNN & 0.96 \std{0.01} \\ \cline{2-4}
        & \multirow{2}{*}{76to100} & GNN & 0.95 \std{0.02} \\ 
         &  & EG-GNN & 0.98 \std{0.01} \\ \midrule
        \multirow{8}{*}{TPSA} & \multirow{2}{*}{0to25} & GNN & 0.97 \std{0.02} \\ 
         &  & EG-GNN & 0.94 \std{0.04} \\ \cline{2-4}
         & \multirow{2}{*}{26to50} & GNN & 0.95 \std{0.02} \\ 
         &  & EG-GNN & 0.99 \std{0.01} \\ \cline{2-4}
         & \multirow{2}{*}{51to75} & GNN & 0.93 \std{0.03} \\ 
         &  & EG-GNN & 0.94 \std{0.04} \\ \cline{2-4}
         & \multirow{2}{*}{76to100} & GNN & 0.94 \std{0.01} \\ 
         &  & EG-GNN & 0.98 \std{0.01} \\
         \bottomrule
    \end{tabular}
    
    \caption{\textbf{Performance results for \mutag using ground-truth explanations in EG-GNNs using GAT.} ~\looseness=-1 Comparison of baseline GNNs and EG-GNNs across OOD test splits and molecular properties. The table reports IID balanced accuracy (IID BACC). Results are presented as mean $\pm$ standard deviation across cross-validation folds for the IID setting.}
    \label{tab:egs_gte_fc}
\end{table}

\begin{table}[H]
    \centering
    \scriptsize
    \begin{tabular}{lccc}
    \toprule
        Property Type & OOD Testset & Model & IID BACC \\ \midrule
        \multirow{8}{*}{MolWt} & \multirow{2}{*}{0to25} & GNN & 0.98 \std{0.01} \\ 
        ~ &  & EG-GNN & 0.98 \std{0.01} \\ \cline{2-4}
        ~ & \multirow{2}{*}{26to50} & GNN & 0.96 \std{0.02} \\ 
        ~ &  & EG-GNN & 0.97 \std{0.01} \\ \cline{2-4}
        ~ & \multirow{2}{*}{51to75} & GNN & 0.96 \std{0.02} \\ 
        ~ &  & EG-GNN & 0.97 \std{0.01} \\ \cline{2-4}
        ~ & \multirow{2}{*}{76to100} & GNN & 0.95 \std{0.03} \\ 
        ~ &  & EG-GNN & 0.95 \std{0.01} \\ \midrule
        \multirow{8}{*}{FractionCSP3} & \multirow{2}{*}{0to25} & GNN & 0.97 \std{0.01} \\ 
        &  & EG-GNN & 0.98 \std{0.01} \\ \cline{2-4}
         & \multirow{2}{*}{26to50} & GNN & 0.96 \std{0.02} \\ 
         &  & EG-GNN & 0.97 \std{0.01} \\ \cline{2-4}
         & \multirow{2}{*}{51to75} & GNN & 0.97 \std{0.01} \\ 
         &  & EG-GNN & 0.98 \std{0.01} \\ \cline{2-4}
         & \multirow{2}{*}{76to100} & GNN & 0.98 \std{0.01} \\ 
         &  & EG-GNN & 0.97 \std{0.01} \\ \midrule
        \multirow{8}{*}{NumRB} & \multirow{2}{*}{0to25} & GNN & 0.98 \std{0.01} \\ 
         &  & EG-GNN & 0.97 \std{0.01} \\ \cline{2-4}
         & \multirow{2}{*}{26to50} & GNN & 0.97 \std{0.01} \\ 
         &  & EG-GNN & 0.97 \std{0.01} \\ \cline{2-4}
         & \multirow{2}{*}{51to75} & GNN & 0.98 \std{0.01} \\ 
         &  & EG-GNN & 0.98 \std{0.01} \\ \cline{2-4}
        & \multirow{2}{*}{76to100} & GNN & 0.97 \std{0.02} \\ 
         &  & EG-GNN & 0.98 \std{0.02} \\ \midrule
        \multirow{8}{*}{TPSA} & \multirow{2}{*}{0to25} & GNN & 0.98 \std{0.01} \\ 
         &  & EG-GNN & 0.97 \std{0.01} \\ \cline{2-4}
         & \multirow{2}{*}{26to50} & GNN & 0.97 \std{0.01} \\ 
         &  & EG-GNN & 0.97 \std{0.01} \\ \cline{2-4}
         & \multirow{2}{*}{51to75} & GNN & 0.96 \std{0.01} \\ 
         &  & EG-GNN & 0.96 \std{0.01} \\ \cline{2-4}
         & \multirow{2}{*}{76to100} & GNN & 0.97 \std{0.01} \\ 
         &  & EG-GNN & 0.97 \std{0.01} \\
         \bottomrule
    \end{tabular}
    
    \caption{\textbf{Performance results for \mutag using ground-truth explanations in EG-GNNs using SAGE.} ~\looseness=-1 Comparison of baseline GNNs and EG-GNNs across OOD test splits and molecular properties. The table reports IID balanced accuracy (IID BACC). Results are presented as mean $\pm$ standard deviation across cross-validation folds for the IID setting.}
    \label{tab:egs_gte_fc}
\end{table}

\begin{table}[H]
    \centering
    \scriptsize
    \begin{tabular}{lccc}
    \toprule
        Property Type & OOD Testset & Model & IID BACC \\ \midrule
        \multirow{8}{*}{MolWt} & \multirow{2}{*}{0to25} & GNN & 0.94 \std{0.05} \\ 
        ~ &  & EG-GNN & 0.94 \std{0.05} \\ \cline{2-4}
        ~ & \multirow{2}{*}{26to50} & GNN & 0.96 \std{0.02} \\ 
        ~ &  & EG-GNN & 0.95 \std{0.02} \\ \cline{2-4}
        ~ & \multirow{2}{*}{51to75} & GNN & 0.96 \std{0.01} \\ 
        ~ &  & EG-GNN & 0.96 \std{0.02} \\ \cline{2-4}
        ~ & \multirow{2}{*}{76to100} & GNN & 0.94 \std{0.02} \\ 
        ~ &  & EG-GNN & 0.95 \std{0.01} \\ \midrule
        \multirow{8}{*}{FractionCSP3} & \multirow{2}{*}{0to25} & GNN & 0.97 \std{0.02} \\ 
        &  & EG-GNN & 0.97 \std{0.01} \\ \cline{2-4}
         & \multirow{2}{*}{26to50} & GNN & 0.95 \std{0.04} \\ 
         &  & EG-GNN & 0.95 \std{0.02} \\ \cline{2-4}
         & \multirow{2}{*}{51to75} & GNN & 0.98 \std{0.01} \\ 
         &  & EG-GNN & 0.97 \std{0.02} \\ \cline{2-4}
         & \multirow{2}{*}{76to100} & GNN & 0.97 \std{0.01} \\ 
         &  & EG-GNN & 0.97 \std{0.01} \\ \midrule
        \multirow{8}{*}{NumRB} & \multirow{2}{*}{0to25} & GNN & 0.97 \std{0.01} \\ 
         &  & EG-GNN & 0.97 \std{0.01} \\ \cline{2-4}
         & \multirow{2}{*}{26to50} & GNN & 0.95 \std{0.02} \\ 
         &  & EG-GNN & 0.96 \std{0.02} \\ \cline{2-4}
         & \multirow{2}{*}{51to75} & GNN & 0.98 \std{0.01} \\ 
         &  & EG-GNN & 0.97 \std{0.02} \\ \cline{2-4}
        & \multirow{2}{*}{76to100} & GNN & 0.97 \std{0.02} \\ 
         &  & EG-GNN & 0.98 \std{0.01} \\ \midrule
        \multirow{8}{*}{TPSA} & \multirow{2}{*}{0to25} & GNN & 0.97 \std{0.01} \\ 
         &  & EG-GNN & 0.98 \std{0.01} \\ \cline{2-4}
         & \multirow{2}{*}{26to50} & GNN & 0.98 \std{0.01} \\ 
         &  & EG-GNN & 0.97 \std{0.01} \\ \cline{2-4}
         & \multirow{2}{*}{51to75} & GNN & 0.97 \std{0.01} \\ 
         &  & EG-GNN & 0.93 \std{0.07} \\ \cline{2-4}
         & \multirow{2}{*}{76to100} & GNN & 0.97 \std{0.02} \\ 
         &  & EG-GNN & 0.98 \std{0.02} \\
         \bottomrule
    \end{tabular}
    %
    \caption{\textbf{Performance results for \mutag using ground-truth explanations in EG-GNNs using GIN.} ~\looseness=-1 Comparison of baseline GNNs and EG-GNNs across OOD test splits and molecular properties. The table reports IID balanced accuracy (IID BACC). Results are presented as mean $\pm$ standard deviation across cross-validation folds for the IID setting.}
    \label{tab:egs_gte_fc}
\end{table}

\begin{table}[H]
    \centering\scriptsize
    \begin{tabular}{lccc}
    \toprule
        Property Type & OOD Testset & Model & IID BACC \\ \midrule
        \multirow{8}{*}{ProbConn} & \multirow{2}{*}{0to25} & GNN & 0.88 \std{0.05} \\ 
        ~ &  & EG-GNN & 0.87 \std{0.06} \\ \cline{2-4}
        ~ & \multirow{2}{*}{26to50} & GNN & 0.95 \std{0.03} \\ 
        ~ &  & EG-GNN & 0.95 \std{0.03} \\ \cline{2-4}
        ~ & \multirow{2}{*}{51to75} & GNN & 0.98 \std{0.01} \\ 
        ~ &  & EG-GNN & 0.98 \std{0.02} \\ \cline{2-4}
        ~ & \multirow{2}{*}{76to100} & GNN & 0.97 \std{0.01} \\ 
        ~ &  & EG-GNN & 0.97 \std{0.02} \\ \midrule
        \multirow{8}{*}{NumSubgraph} & \multirow{2}{*}{0to25} & GNN & 0.85 \std{0.04} \\ 
        &  & EG-GNN & 0.85 \std{0.05} \\ \cline{2-4}
         & \multirow{2}{*}{26to50} & GNN & 0.93 \std{0.03} \\ 
         &  & EG-GNN & 0.94 \std{0.02} \\ \cline{2-4}
         & \multirow{2}{*}{51to75} & GNN & 0.98 \std{0.01} \\ 
         &  & EG-GNN & 0.98 \std{0.01} \\ \cline{2-4}
         & \multirow{2}{*}{76to100} & GNN & 0.78 \std{0.06} \\ 
         &  & EG-GNN & 0.74 \std{0.06} \\ \midrule
        \multirow{8}{*}{SubgraphSize} & \multirow{2}{*}{0to25} & GNN & 0.94 \std{0.04} \\ 
         &  & EG-GNN & 0.95 \std{0.04} \\ \cline{2-4}
         & \multirow{2}{*}{26to50} & GNN & 0.94 \std{0.02} \\ 
         &  & EG-GNN & 0.95 \std{0.03} \\ \cline{2-4}
         & \multirow{2}{*}{51to75} & GNN & 0.90 \std{0.01} \\ 
         &  & EG-GNN & 0.90 \std{0.01} \\ \cline{2-4}
        & \multirow{2}{*}{76to100} & GNN & 0.89 \std{0.05} \\ 
         &  & EG-GNN & 0.87 \std{0.05} \\ \midrule
    \end{tabular}
    
    \caption{\textbf{Performance results for \tri using ground-truth explanations in EG-GNNs using GCN.} ~\looseness=-1 Comparison of baseline GNNs and EG-GNNs across OOD test splits and graph properties. The table reports IID balanced accuracy (IID BACC). Results are presented as mean $\pm$ standard deviation across cross-validation folds for the IID setting.}
    \label{tab:egs_gte_fc}
\end{table}

\begin{table}[H]
    \centering\scriptsize
    \begin{tabular}{lccc}
    \toprule
        Property Type & OOD Testset & Model & IID BACC \\ \midrule
        \multirow{8}{*}{ProbConn} & \multirow{2}{*}{0to25} & GNN & 0.74 \std{0.06} \\ 
        ~ &  & EG-GNN & 0.71 \std{0.01} \\ \cline{2-4}
        ~ & \multirow{2}{*}{26to50} & GNN & 0.90 \std{0.03} \\ 
        ~ &  & EG-GNN & 0.89 \std{0.05} \\ \cline{2-4}
        ~ & \multirow{2}{*}{51to75} & GNN & 0.91 \std{0.03} \\ 
        ~ &  & EG-GNN & 0.93 \std{0.02} \\ \cline{2-4}
        ~ & \multirow{2}{*}{76to100} & GNN & 0.95 \std{0.03} \\ 
        ~ &  & EG-GNN & 0.93 \std{0.05} \\ \midrule
        \multirow{8}{*}{NumSubgraph} & \multirow{2}{*}{0to25} & GNN & 0.76 \std{0.04} \\ 
        &  & EG-GNN & 0.77 \std{0.07} \\ \cline{2-4}
         & \multirow{2}{*}{26to50} & GNN & 0.89 \std{0.03} \\ 
         &  & EG-GNN & 0.87 \std{0.03} \\ \cline{2-4}
         & \multirow{2}{*}{51to75} & GNN & 0.98 \std{0.02} \\ 
         &  & EG-GNN & 0.98 \std{0.01} \\ \cline{2-4}
         & \multirow{2}{*}{76to100} & GNN & 0.69 \std{0.06} \\ 
         &  & EG-GNN & 0.59 \std{0.10} \\ \midrule
        \multirow{8}{*}{SubgraphSize} & \multirow{2}{*}{0to25} & GNN & 0.87 \std{0.04} \\ 
         &  & EG-GNN & 0.92 \std{0.04} \\ \cline{2-4}
         & \multirow{2}{*}{26to50} & GNN & 0.86 \std{0.05} \\ 
         &  & EG-GNN & 0.89 \std{0.03} \\ \cline{2-4}
         & \multirow{2}{*}{51to75} & GNN & 0.96 \std{0.02} \\ 
         &  & EG-GNN & 0.97 \std{0.03} \\ \cline{2-4}
        & \multirow{2}{*}{76to100} & GNN & 0.96 \std{0.01} \\ 
         &  & EG-GNN & 0.96 \std{0.03} \\ \midrule
    \end{tabular}
    
    \caption{\textbf{Performance results for \tri using ground-truth explanations in EG-GNNs using GAT.} ~\looseness=-1 Comparison of baseline GNNs and EG-GNNs across OOD test splits and graph properties. The table reports IID balanced accuracy (IID BACC). Results are presented as mean $\pm$ standard deviation across cross-validation folds for the IID setting.}
    \label{tab:egs_gte_fc}
\end{table}

\begin{table}[H]
    \centering\scriptsize
    \begin{tabular}{lccc}
    \toprule
        Property Type & OOD Testset & Model & IID BACC \\ \midrule
        \multirow{8}{*}{ProbConn} & \multirow{2}{*}{0to25} & GNN & 0.86 \std{0.06} \\ 
        ~ &  & EG-GNN & 0.90 \std{0.03} \\ \cline{2-4}
        ~ & \multirow{2}{*}{26to50} & GNN & 0.90 \std{0.03} \\ 
        ~ &  & EG-GNN & 0.92 \std{0.03} \\ \cline{2-4}
        ~ & \multirow{2}{*}{51to75} & GNN & 0.97 \std{0.01} \\ 
        ~ &  & EG-GNN & 0.96 \std{0.03} \\ \cline{2-4}
        ~ & \multirow{2}{*}{76to100} & GNN & 0.95 \std{0.04} \\ 
        ~ &  & EG-GNN & 0.95 \std{0.05} \\ \midrule
        \multirow{8}{*}{NumSubgraph} & \multirow{2}{*}{0to25} & GNN & 0.88 \std{0.04} \\ 
        &  & EG-GNN & 0.90 \std{0.03} \\ \cline{2-4}
         & \multirow{2}{*}{26to50} & GNN & 0.93 \std{0.04} \\ 
         &  & EG-GNN & 0.93 \std{0.04} \\ \cline{2-4}
         & \multirow{2}{*}{51to75} & GNN & 0.97 \std{0.01} \\ 
         &  & EG-GNN & 0.96 \std{0.03} \\ \cline{2-4}
         & \multirow{2}{*}{76to100} & GNN & 0.60 \std{0.07} \\ 
         &  & EG-GNN & 0.61 \std{0.07} \\ \midrule
        \multirow{8}{*}{SubgraphSize} & \multirow{2}{*}{0to25} & GNN & 0.91 \std{0.06} \\ 
         &  & EG-GNN & 0.91 \std{0.03} \\ \cline{2-4}
         & \multirow{2}{*}{26to50} & GNN & 0.92 \std{0.02} \\ 
         &  & EG-GNN & 0.94 \std{0.01} \\ \cline{2-4}
         & \multirow{2}{*}{51to75} & GNN & 0.90 \std{0.02} \\ 
         &  & EG-GNN & 0.90 \std{0.01} \\ \cline{2-4}
        & \multirow{2}{*}{76to100} & GNN & 0.92 \std{0.04} \\ 
         &  & EG-GNN & 0.94 \std{0.05} \\ \midrule
    \end{tabular}
    
    \caption{\textbf{Performance results for \tri using ground-truth explanations in EG-GNNs using SAGE.} ~\looseness=-1 Comparison of baseline GNNs and EG-GNNs across OOD test splits and graph properties. The table reports IID balanced accuracy (IID BACC). Results are presented as mean $\pm$ standard deviation across cross-validation folds for the IID setting.}
    \label{tab:egs_gte_fc}
\end{table}

\begin{table}[H]
    \centering\scriptsize
    \begin{tabular}{lccc}
    \toprule
        Property Type & OOD Testset & Model & IID BACC \\ \midrule
        \multirow{8}{*}{ProbConn} & \multirow{2}{*}{0to25} & GNN & 0.95 \std{0.03} \\ 
        ~ &  & EG-GNN & 0.95 \std{0.03} \\ \cline{2-4}
        ~ & \multirow{2}{*}{26to50} & GNN & 0.98 \std{0.01} \\ 
        ~ &  & EG-GNN & 0.98 \std{0.01} \\ \cline{2-4}
        ~ & \multirow{2}{*}{51to75} & GNN & 0.98 \std{0.01} \\ 
        ~ &  & EG-GNN & 0.98 \std{0.01} \\ \cline{2-4}
        ~ & \multirow{2}{*}{76to100} & GNN & 0.99 \std{0.02} \\ 
        ~ &  & EG-GNN & 0.98 \std{0.01} \\ \midrule
        \multirow{8}{*}{NumSubgraph} & \multirow{2}{*}{0to25} & GNN & 0.93 \std{0.03} \\ 
        &  & EG-GNN & 0.93 \std{0.03} \\ \cline{2-4}
         & \multirow{2}{*}{26to50} & GNN & 0.98 \std{0.02} \\ 
         &  & EG-GNN & 0.98 \std{0.02} \\ \cline{2-4}
         & \multirow{2}{*}{51to75} & GNN & 0.98 \std{0.01} \\ 
         &  & EG-GNN & 0.98 \std{0.01} \\ \cline{2-4}
         & \multirow{2}{*}{76to100} & GNN & 0.99 \std{0.01} \\ 
         &  & EG-GNN & 1.00 \std{0.01} \\ \midrule
        \multirow{8}{*}{SubgraphSize} & \multirow{2}{*}{0to25} & GNN & 0.99 \std{0.01} \\ 
         &  & EG-GNN & 0.99 \std{0.01} \\ \cline{2-4}
         & \multirow{2}{*}{26to50} & GNN & 0.97 \std{0.02} \\ 
         &  & EG-GNN & 0.97 \std{0.03} \\ \cline{2-4}
         & \multirow{2}{*}{51to75} & GNN & 0.95 \std{0.03} \\ 
         &  & EG-GNN & 0.96 \std{0.02} \\ \cline{2-4}
        & \multirow{2}{*}{76to100} & GNN & 0.92 \std{0.03} \\ 
         &  & EG-GNN & 0.94 \std{0.02} \\ \midrule
    \end{tabular}
    
    \caption{\textbf{Performance results for \tri using ground-truth explanations in EG-GNNs using GIN.} ~\looseness=-1 Comparison of baseline GNNs and EG-GNNs across OOD test splits and graph properties. The table reports IID balanced accuracy (IID BACC). Results are presented as mean $\pm$ standard deviation across cross-validation folds for the IID setting.}
    \label{tab:egs_gte_fc}
\end{table}

\begin{table}[H]
    \centering\scriptsize
    \begin{tabular}{lccc}
    \toprule
        Property Type & OOD Testset & Model & IID BACC \\ \midrule
        \multirow{8}{*}{ProbConn} & \multirow{2}{*}{0to25} & GNN & 0.81 \std{0.06} \\ 
        ~ &  & EG-GNN & 0.82 \std{0.06} \\ \cline{2-4}
        ~ & \multirow{2}{*}{26to50} & GNN & 0.94 \std{0.02} \\ 
        ~ &  & EG-GNN & 0.90 \std{0.02} \\ \cline{2-4}
        ~ & \multirow{2}{*}{51to75} & GNN & 0.96 \std{0.02} \\ 
        ~ &  & EG-GNN & 0.93 \std{0.05} \\ \cline{2-4}
        ~ & \multirow{2}{*}{76to100} & GNN & 0.95 \std{0.03} \\ 
        ~ &  & EG-GNN & 0.90 \std{0.03} \\ \midrule
        \multirow{8}{*}{NumSubgraph} & \multirow{2}{*}{0to25} & GNN & 0.83 \std{0.03} \\ 
        &  & EG-GNN & 0.84 \std{0.02} \\ \cline{2-4}
         & \multirow{2}{*}{26to50} & GNN & 0.94 \std{0.04} \\ 
         &  & EG-GNN & 0.89 \std{0.05} \\ \cline{2-4}
         & \multirow{2}{*}{51to75} & GNN & 0.94 \std{0.02} \\ 
         &  & EG-GNN & 0.86 \std{0.07} \\ \cline{2-4}
         & \multirow{2}{*}{76to100} & GNN & 0.85 \std{0.05} \\ 
         &  & EG-GNN & 0.81 \std{0.06} \\ \midrule
        \multirow{8}{*}{SubgraphSize} & \multirow{2}{*}{0to25} & GNN & 0.98 \std{0.02} \\ 
         &  & EG-GNN & 0.97 \std{0.01} \\ \cline{2-4}
         & \multirow{2}{*}{26to50} & GNN & 0.96 \std{0.03} \\ 
         &  & EG-GNN & 0.94 \std{0.02} \\ \cline{2-4}
         & \multirow{2}{*}{51to75} & GNN & 0.90 \std{0.06} \\ 
         &  & EG-GNN & 0.85 \std{0.04} \\ \cline{2-4}
        & \multirow{2}{*}{76to100} & GNN & 0.85 \std{0.04} \\ 
         &  & EG-GNN & 0.81 \std{0.02} \\ \midrule
    \end{tabular}
    
    \caption{\textbf{Performance results for \pen using ground-truth explanations in EG-GNNs using GCN.} ~\looseness=-1 Comparison of baseline GNNs and EG-GNNs across OOD test splits and graph properties. The table reports IID balanced accuracy (IID BACC). Results are presented as mean $\pm$ standard deviation across cross-validation folds for the IID setting.}
    \label{tab:egs_gte_fc}
\end{table}

\begin{table}[H]
    \centering\scriptsize
    \begin{tabular}{lccc}
    \toprule
        Property Type & OOD Testset & Model & IID BACC \\ \midrule
        \multirow{8}{*}{ProbConn} & \multirow{2}{*}{0to25} & GNN & 0.79 \std{0.05} \\ 
        ~ &  & EG-GNN & 0.82 \std{0.05} \\ \cline{2-4}
        ~ & \multirow{2}{*}{26to50} & GNN & 0.96 \std{0.03} \\ 
        ~ &  & EG-GNN & 0.97 \std{0.02} \\ \cline{2-4}
        ~ & \multirow{2}{*}{51to75} & GNN & 0.89 \std{0.03} \\ 
        ~ &  & EG-GNN & 0.91 \std{0.04} \\ \cline{2-4}
        ~ & \multirow{2}{*}{76to100} & GNN & 0.91 \std{0.04} \\ 
        ~ &  & EG-GNN & 0.92 \std{0.04} \\ \midrule
        \multirow{8}{*}{NumSubgraph} & \multirow{2}{*}{0to25} & GNN & 0.79 \std{0.02} \\ 
        &  & EG-GNN & 0.78 \std{0.04} \\ \cline{2-4}
         & \multirow{2}{*}{26to50} & GNN & 0.92 \std{0.04} \\ 
         &  & EG-GNN & 0.94 \std{0.02} \\ \cline{2-4}
         & \multirow{2}{*}{51to75} & GNN & 0.90 \std{0.05} \\ 
         &  & EG-GNN & 0.91 \std{0.02} \\ \cline{2-4}
         & \multirow{2}{*}{76to100} & GNN & 0.84 \std{0.03} \\ 
         &  & EG-GNN & 0.80 \std{0.08} \\ \midrule
        \multirow{8}{*}{SubgraphSize} & \multirow{2}{*}{0to25} & GNN & 0.92 \std{0.03} \\ 
         &  & EG-GNN & 0.91 \std{0.03} \\ \cline{2-4}
         & \multirow{2}{*}{26to50} & GNN & 0.93 \std{0.05} \\ 
         &  & EG-GNN & 0.92 \std{0.03} \\ \cline{2-4}
         & \multirow{2}{*}{51to75} & GNN & 0.98 \std{0.02} \\ 
         &  & EG-GNN & 0.98 \std{0.02} \\ \cline{2-4}
        & \multirow{2}{*}{76to100} & GNN & 0.99 \std{0.02} \\ 
         &  & EG-GNN & 0.97 \std{0.02} \\ \midrule
    \end{tabular}
    
    \caption{\textbf{Performance results for \pen using ground-truth explanations in EG-GNNs using GAT.} ~\looseness=-1 Comparison of baseline GNNs and EG-GNNs across OOD test splits and graph properties. The table reports IID balanced accuracy (IID BACC). Results are presented as mean $\pm$ standard deviation across cross-validation folds for the IID setting.}
    \label{tab:egs_gte_fc}
\end{table}

\begin{table}[H]
    \centering\scriptsize
    \begin{tabular}{lccc}
    \toprule
        Property Type & OOD Testset & Model & IID BACC \\ \midrule
        \multirow{8}{*}{ProbConn} & \multirow{2}{*}{0to25} & GNN & 0.81 \std{0.07} \\ 
        ~ &  & EG-GNN & 0.80 \std{0.09} \\ \cline{2-4}
        ~ & \multirow{2}{*}{26to50} & GNN & 0.90 \std{0.03} \\ 
        ~ &  & EG-GNN & 0.86 \std{0.05} \\ \cline{2-4}
        ~ & \multirow{2}{*}{51to75} & GNN & 0.85 \std{0.07} \\ 
        ~ &  & EG-GNN & 0.85 \std{0.06} \\ \cline{2-4}
        ~ & \multirow{2}{*}{76to100} & GNN & 0.87 \std{0.04} \\ 
        ~ &  & EG-GNN & 0.89 \std{0.04} \\ \midrule
        \multirow{8}{*}{NumSubgraph} & \multirow{2}{*}{0to25} & GNN & 0.87 \std{0.04} \\ 
        &  & EG-GNN & 0.88 \std{0.04} \\ \cline{2-4}
         & \multirow{2}{*}{26to50} & GNN & 0.86 \std{0.03} \\ 
         &  & EG-GNN & 0.83 \std{0.03} \\ \cline{2-4}
         & \multirow{2}{*}{51to75} & GNN & 0.74 \std{0.06} \\ 
         &  & EG-GNN & 0.74 \std{0.06} \\ \cline{2-4}
         & \multirow{2}{*}{76to100} & GNN & 0.64 \std{0.05} \\ 
         &  & EG-GNN & 0.65 \std{0.05} \\ \midrule
        \multirow{8}{*}{SubgraphSize} & \multirow{2}{*}{0to25} & GNN & 0.96 \std{0.02} \\ 
         &  & EG-GNN & 0.97 \std{0.03} \\ \cline{2-4}
         & \multirow{2}{*}{26to50} & GNN & 0.90 \std{0.05} \\ 
         &  & EG-GNN & 0.89 \std{0.06} \\ \cline{2-4}
         & \multirow{2}{*}{51to75} & GNN & 0.81 \std{0.05} \\ 
         &  & EG-GNN & 0.82 \std{0.05} \\ \cline{2-4}
        & \multirow{2}{*}{76to100} & GNN & 0.74 \std{0.05} \\ 
         &  & EG-GNN & 0.75 \std{0.05} \\ \midrule
    \end{tabular}
    
    \caption{\textbf{Performance results for \pen using ground-truth explanations in EG-GNNs using SAGE.} ~\looseness=-1 Comparison of baseline GNNs and EG-GNNs across OOD test splits and graph properties. The table reports IID balanced accuracy (IID BACC). Results are presented as mean $\pm$ standard deviation across cross-validation folds for the IID setting.}
    \label{tab:egs_gte_fc}
\end{table}

\begin{table}[H]
    \centering\scriptsize
    \begin{tabular}{lccc}
    \toprule
        Property Type & OOD Testset & Model & IID BACC \\ \midrule
        \multirow{8}{*}{ProbConn} & \multirow{2}{*}{0to25} & GNN & 0.76 \std{0.04} \\ 
        ~ &  & EG-GNN & 0.74 \std{0.04} \\ \cline{2-4}
        ~ & \multirow{2}{*}{26to50} & GNN & 0.88 \std{0.06} \\ 
        ~ &  & EG-GNN & 0.89 \std{0.04} \\ \cline{2-4}
        ~ & \multirow{2}{*}{51to75} & GNN & 0.90 \std{0.03} \\ 
        ~ &  & EG-GNN & 0.90 \std{0.04} \\ \cline{2-4}
        ~ & \multirow{2}{*}{76to100} & GNN & 0.81 \std{0.04} \\ 
        ~ &  & EG-GNN & 0.83 \std{0.05} \\ \midrule
        \multirow{8}{*}{NumSubgraph} & \multirow{2}{*}{0to25} & GNN & 0.82 \std{0.01} \\ 
        &  & EG-GNN & 0.82 \std{0.03} \\ \cline{2-4}
         & \multirow{2}{*}{26to50} & GNN & 0.84 \std{0.03} \\ 
         &  & EG-GNN & 0.86 \std{0.04} \\ \cline{2-4}
         & \multirow{2}{*}{51to75} & GNN & 0.89 \std{0.05} \\ 
         &  & EG-GNN & 0.82 \std{0.03} \\ \cline{2-4}
         & \multirow{2}{*}{76to100} & GNN & 0.87 \std{0.02} \\ 
         &  & EG-GNN & 0.90 \std{0.03} \\ \midrule
        \multirow{8}{*}{SubgraphSize} & \multirow{2}{*}{0to25} & GNN & 0.89 \std{0.02} \\ 
         &  & EG-GNN & 0.90 \std{0.02} \\ \cline{2-4}
         & \multirow{2}{*}{26to50} & GNN & 0.84 \std{0.03} \\ 
         &  & EG-GNN & 0.82 \std{0.05} \\ \cline{2-4}
         & \multirow{2}{*}{51to75} & GNN & 0.90 \std{0.02} \\ 
         &  & EG-GNN & 0.92 \std{0.01} \\ \cline{2-4}
        & \multirow{2}{*}{76to100} & GNN & 0.86 \std{0.02} \\ 
         &  & EG-GNN & 0.88 \std{0.04} \\ \midrule
    \end{tabular}
    
    \caption{\textbf{Performance results for \pen using ground-truth explanations in EG-GNNs using GIN.} ~\looseness=-1 Comparison of baseline GNNs and EG-GNNs across OOD test splits and graph properties. The table reports IID balanced accuracy (IID BACC). Results are presented as mean $\pm$ standard deviation across cross-validation folds for the IID setting.}
    \label{tab:egs_gte_fc}
\end{table}

\end{document}